\documentclass[10pt,conference]{IEEEtran}
\usepackage{spconf,amsmath,amsfonts,graphicx}
\usepackage[numbers,sort&compress]{natbib}
\usepackage{algorithmic}
\usepackage{graphicx}
\usepackage{textcomp}
\usepackage{xcolor}
\usepackage{soul}
\usepackage{bbold}
\usepackage{algorithm}
\usepackage{tabularx}
\usepackage{adjustbox}
\usepackage{booktabs}
\usepackage{subcaption}

\newcommand{\Ls}{\mathcal{L}}
\newcommand{\T}{\mathcal{T}}
\newcommand{\D}{\mathcal{D}}
\newcommand{\F}{\mathcal{F}}
\newcommand{\St}{\mathcal{S}}

\newif\iffinal
\finalfalse

\iffinal
\else
\usepackage[switch]{lineno}
\fi

\begin{document}

\title{Learning to Segment Medical Images from Few-Shot Sparse Labels}
%
%
%
%
\name{Pedro H. T. Gama$^{\star}$ \qquad Hugo Oliveira$^{\star\dagger}$ \qquad Jefersson A. dos Santos$^{\star}$
\thanks{\textbf{Acknowledgments:} We would like to thanks the CNPq, CAPES, Fapemig, Fapesp, for partially funding this research, and NVIDIA for the donation of GPUs to the PATREO Laboratory that were used in this work.}
}
\address{$^{\star}$ Department of Computer Science, Universidade Federal de Minas Gerais, Brazil \\ $^{\dagger}$ Institute of Mathematics and Statistics, University of S\~{a}o Paulo, Brazil}

\maketitle

\begin{abstract}
In this paper, we propose a novel approach for few-shot semantic segmentation with sparse labeled images.
We investigate the effectiveness of our method, which is based on the Model-Agnostic Meta-Learning (MAML) algorithm, in the medical scenario, where the use of sparse labeling and few-shot can alleviate the cost of producing new annotated datasets. Our method uses sparse labels in the meta-training and dense labels in the meta-test, thus making the model learn to predict dense labels from sparse ones. We conducted experiments with four Chest X-Ray datasets to evaluate two types of annotations (grid and points). The results show that our method is the most suitable when the target domain highly differs from source domains, achieving Jaccard scores comparable to dense labels, using less than $2\%$ of the pixels of an image with labels in few-shot scenarios.
\end{abstract}


\IEEEpeerreviewmaketitle



\newcommand{\currprop}{0.9\columnwidth}

\section{Introduction}
\label{sec:introduction}


Medical images are useful tools to assist doctors in multiple clinical scenarios and to plan for surgery. X-Ray, Magnetic Resonance Imaging (MRI), Computed Tomography (CT) and other imaging modalities are non-invasive methods that 
can help in diagnosis, pathology localization, anatomical studies, and other tasks~\cite{sharma2010automated}.

Convolutional Neural Networks (CNNs) and their variants are the state of the art for object classification, detection, semantic segmentation and other Computer Vision problems. Classical convolutional networks are known for their large data requirements, often hindering their usage in scenarios were data availability is limited, as in medical imaging. Relatively few medical datasets are publicly available due to privacy and ethical concerns \cite{shiraishi2000development,jaeger2014two,NIHwang2017chestx,MIASsuckling1994mammographic,IVISIONsilva2018automatic,PANORAMICabdi2015automatic,moreira2012inbreast}.

Even among the public datasets, properly curated labeled data is limited due to the need for specialized annotators (i.e. radiologists), severely hampering the creation of general models for medical image understanding. While many datasets contain image-level annotations indicating the presence or absence of a set of medical conditions, the creation of pixel-level labels that allow for the training of semantic segmentation models is much more laborious. Volumetric image modalities as MRIs or CT scans further compound these difficulties by requiring per-slice annotations, often followed by cross-axes analysis to detect inconsistencies, which can take hours for one single exam. Hence, there is a need for automatic and semi-automatic segmentation methods to assist physicians in the annotation of these images. One way to alleviate the burden of medical professionals in labeling the exams is to improve the generalization capabilities of existing pretrained models. For instance, domain adaptation can be used to transfer knowledge from related medical imaging datasets to improve segmentation performance in unseen target tasks. 
Scenarios with low amounts of data available, often called Few-shot, have been studied in recent years. Tasks such as few-shot classification~\cite{snell2017prototypical,finn2017model} are the most explored in the literature with substantial results for datasets such as MNIST \cite{lecun1998gradient} or Omniglot \cite{lake2015human}. As for the problem of pixel-level annotations, one efficient option is \textit{sparse labeling}, that is, specifying the labels of a small number of pixels. Methods that can make efficient use of few-shot and sparse labels can solve semantic segmentation medical problems on datasets created in a labor-efficient manner.
Thus, \textbf{our main contribution} is the proposal of a novel approach to few-shot semantic segmentation in medical images from sparse labels. For that, we introduce the \textbf{Wea}kly-supervised \textbf{Se}gmentation \textbf{L}earning (WeaSeL), which extends the MAML~\cite{finn2017model} algorithm by introducing annotation sparsity directly to its meta-training stage.

\section{Related Work}
\label{sec:related}


\noindent\textbf{Biomedical Image Segmentation:} Automatic and semi-automatic segmentation of medical images has been studied for decades, as such methods can considerably alleviate the burden of physicians in labeling these data \cite{ronneberger2015u}. Medical image segmentation from sparse labels can be especially useful, since these images often have a small number of labeled samples due to data privacy restrictions and the lack of specialists for annotating the samples. The survey of Tajbakhsh \textit{et. al} \cite{tajbakhsh2020embracing} reviews a collection of Deep Learning (DL) solutions on medical image segmentation, which includes a section for segmentation with noisy/sparse labels. The methods reviewed can be summarized in a selective loss with or without mask completion. Sparse segmentation methods either use a loss function able to ignore unlabeled pixels/voxels or employ some technique to augment the sparse annotations to resemble dense masks.

\noindent\textbf{Meta-Learning:} Few-shot learning has attracted considerable attention over the last years, mainly due to recent advances in Deep Learning for Self-Supervised learning \cite{goyal2019scaling} and Meta-Learning \cite{snell2017prototypical,finn2017model}. Meta-Learning has become a proliferous research topic in Deep Learning, as the literature aims to improve on the generalization capabilities of Deep Neural Networks (DNNs). Within Meta-Learning, two prominent and distinct methodologies gained attention: Gradient-based and Metric learning. Finn \textit{et al.} \cite{finn2017model} proposed the MAML framework, based on the trend of Gradient-based approaches. MAML uses multiple tasks -- that is, multiple data/sample pairs and a loss function -- during its meta-training to create a generalizable model able to perform quick adaptation and feature reuse to infer over unseen related tasks. 
Meta-Learning arose at first for object classification tasks, while tasks such as detection and segmentation still lacking development. Additionally, the intersection of few-shot learning and medical image segmentation from sparse labels has proven to be quite a challenging task, with very few methods being described in the literature \cite{wang2018interactive}.

\noindent\textbf{Few-shot Semantic Segmentation:} Few-shot segmentation became a relevant topic only recently. Several methods \cite{shaban2017one,dong2018few,hu2019attention,zhang2019sgone,wang2019panet} were recently proposed to make use of the small subset of labeled images, often called \textit{support} set, to segment a query image. However, the vast majority of few-shot segmentation methods in the literature do not explore sparsely labeled images in the support set, only dealing with densely labeled samples. 
Rakelly \textit{et al.} \cite{rakelly2018conditional} proposed the first algorithm for few-shot sparse segmentation: Guided Networks.
Guided Nets use a pretrained CNN backbone to extract features of both the support set and the query image. The weakly labeled support samples and sparse segmentation masks are combined, pooled, and subsequently used as weights to segment the query. The vast majority of few-shot segmentation methods  \cite{shaban2017one,dong2018few,rakelly2018conditional,hu2019attention,zhang2019sgone,wang2019panet} were originally proposed to RGB images as the ones seen in datasets such as the Pascal VOC~\cite{pascal-voc-2012}, thus, most of the literature on this topic rely on pretrained CNN or even Fully Convolutional Networks (FCNs) backbones.

\section{Methodology}
\label{sec:methodology}







We define a segmentation task $\mathcal{S} = \{\D^{sup}, \D^{qry}, c\}$, where $\D$ is a dataset with partitions $\D^{qry}$ (or query set) and $\D^{sup}$ (or support set) such that, $\D^{qry}\cap\D^{sup}=\emptyset$. The class $c$ is the positive/foreground class for the task. A dataset $\D$ is a set of pairs $(\mathbf{x}, \mathbf{y})$, where $\mathbf{x}$ is an image and $\mathbf{y}$ is the respective semantic label, with $\mathbf{y}$ being a dense mask for images in $\D^{qry}$ and a sparsely annotated mask for the $\D^{sup}$ set. In particular, we define a few-shot segmentation task $\mathcal{F}$ as a task wherein $\D^{sup}$ has a small amount of labeled samples (e.g. 20 or less) and $\D^{qry}$ labels are absent or unknown. We refer to a few-shot task as $k$-shot when $k = |\D^{sup}|$, that is, the number of samples in its support set is $k$.

Thus, given a set of segmentation tasks $\mathcal{S} = \{\mathcal{S}_1 , \mathcal{S}_2 , \dots, \mathcal{S}_n \}$, and a target few-shot task $\mathcal{F}$, we want to segment the images from $\D^{qry}_{\mathcal{F}}$ using information from both $\mathcal{S}$ and $\D^{sup}_{\mathcal{F}}$. 
Also, it holds that no pair of image/semantic label of $\F$ is present in any task $\St_i$ in either $qry$ or $sup$ partition.

\subsection{Gradient-based Segmentation from Sparse Labels}
\label{sec:maml_seg}

As previously mentioned, we propose WeaSeL, an gradient-based approach for semantic segmentation derived from MAML~\cite{finn2017model}. A graphical representation of our method can be seen in Figure~\ref{fig:maml_seg}. 

\renewcommand{\currprop}{0.9\columnwidth}

\begin{figure}[!t]
    \centering
    \includegraphics[clip, trim=2.2in 1.5in 2.2in 1.7in, page=1, width=\currprop]{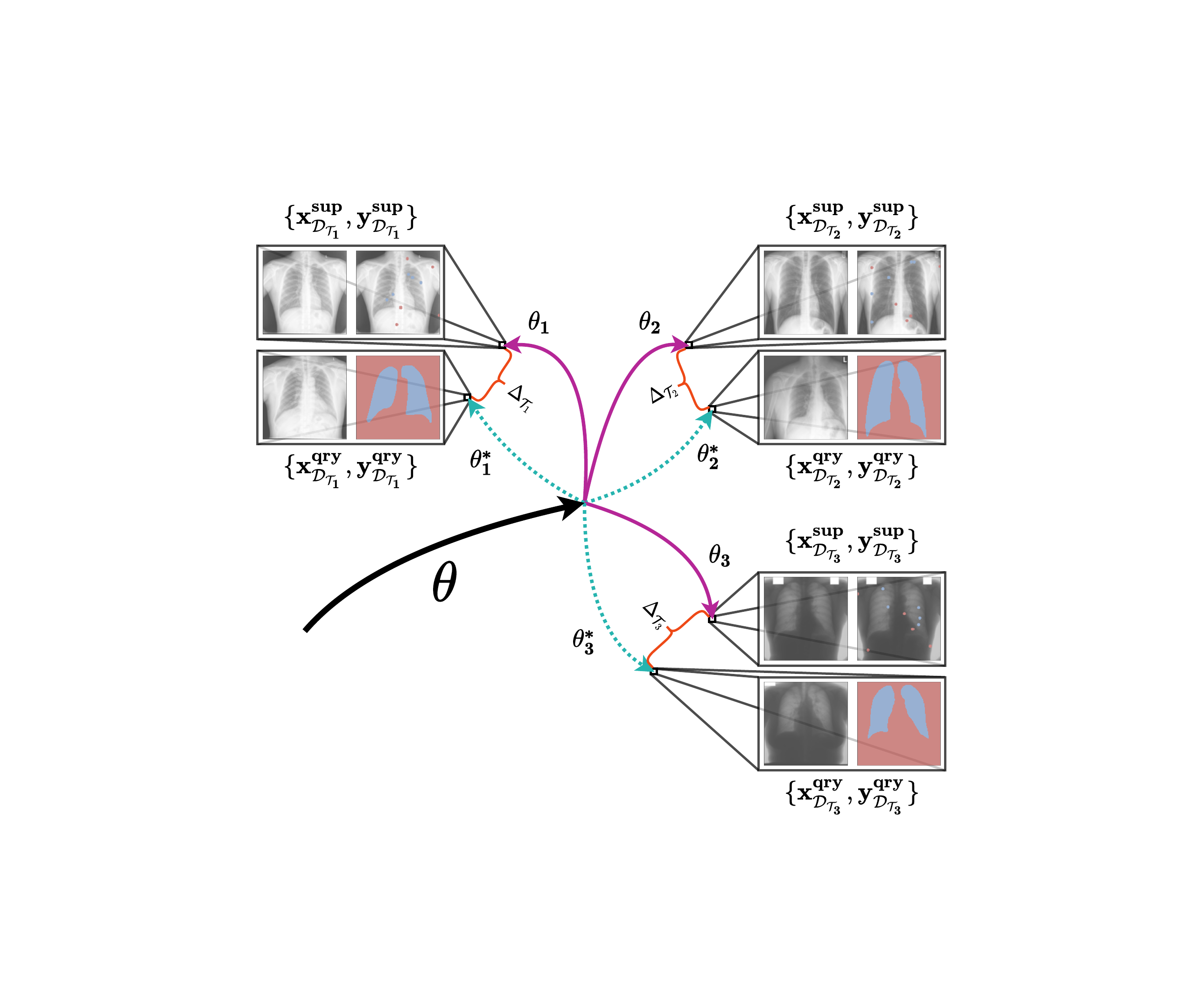}
    \caption{Visualization of the proposed approach. The parameters $\theta_i$ are optimized using sparse labels from support sets. The optimal $\theta_i^*$ would be obtained if dense labels were presented in meta-training, as the ones in the query set used to compute the task outer loss. The model learns to intrinsically minimize this difference $\Delta$ between parameters, and thus fastly adapt to the few-shot task.}
    \label{fig:maml_seg}
\end{figure}

We define a meta task $\T = \{\Ls, \D^{sup}, \D^{qry}, c\}$ comprised of a loss function $\Ls$ and a pair of datasets $\D^{sup}$ and $\D^{qry}$ -- the meta-train and meta-test sets, respectively -- and a class $c$. This is an extension of the segmentation task definition from Section~\ref{sec:methodology}. We assume a distribution over tasks $p(\T)$ that our model $f_\theta$ (parametrized by $\theta$) is desired to be able to adapt to.
A description of this supervised training adaptation of MAML can be seen in Algorithm~\ref{alg:maml_gen}. 


\begin{algorithm}
    \caption{Model-Agnostic Meta Learning: Weakly Supervised}
    \label{alg:maml_gen}
    \begin{algorithmic}
        \REQUIRE $p(\T)$: distributions over tasks
        \REQUIRE $\alpha, \beta$: step size hyperparameters
        
        \STATE Randomly initialize $\theta$
        \WHILE{not done}
        \STATE Sample batch of tasks $\T_i \sim p(\T)$
        \FORALL{$\T_i$}
        \STATE Sample batch of datapoints $S_i = \{(\mathbf{x}, \mathbf{y})\}$ from $\D^{sup}_{\T_i}$
        \STATE Compute $\nabla_\theta\Ls_{\T_i}(f_\theta)$ using $S_i$ and $\Ls_{\T_i}$
        \STATE Update adapted parameters using gradient descent: $\theta_i = \theta - \alpha\nabla_\theta\Ls_{\T_i}(f_\theta)$
        \STATE Sample batch of datapoints $Q_i = \{(\mathbf{x}, \mathbf{y})\}$ from $\D^{qry}_{\T_i}$
        \ENDFOR
        \STATE Update $\theta \leftarrow \theta - \beta\nabla_\theta \sum_{\T_i \sim p(\T)}\Ls_{\T_i}(f_{\theta_i})$ using each $Q_i$ and $\Ls_{\T_i}$
        \ENDWHILE
    \end{algorithmic}
\end{algorithm}

During meta-training, the inner loss is computed using the sparse labels from samples of $\D^{sup}$, while the outer loss component takes into account the dense labels from samples of $\D^{qry}$. This strategy directly encourages the model to generate dense segmentation from the sparse labels fed on the tuning phase. Given that our problem is semantic segmentation, we define the loss function $\Ls_{\T_i}$ for all tasks $\T_i \sim p(\T)$ as the Cross-Entropy loss, ignoring the pixels with unknown labels. This is achieved by computing a weighted Cross-Entropy loss in a pixel-wise fashion, with the caveat that unknown pixels have weight $0$. When the meta-training process is finished, we fine-tune the model on the target $\mathcal{F}$ task, through a conventional supervised training on the $\D_{\mathcal{F}}^{sup}$ set, again, with a weighted cross-entropy loss.

\section{Experimental Setup}
\label{sec:experimental_setup}

\subsection{Datasets}
\label{sec:datasets}

As MAML requires a bundle of tasks to properly learn to learn from few-shot examples, we constructed a Meta-Dataset of radiological image segmentation tasks from publicly available datasets. More specifically, we build this Meta-Dataset using five Chest X-Ray (CXR) \cite{shiraishi2000development,jaeger2014two,pi-null-mezon,NIHwang2017chestx}, two Mammographic X-Ray (MXR) \cite{MIASsuckling1994mammographic,moreira2012inbreast} and two Dental X-Ray (DXR) \cite{PANORAMICabdi2015automatic,IVISIONsilva2018automatic} datasets. Some of these datasets contain segmentation masks for multiple organs, which were all included in the Meta-Dataset.

In Table~\ref{tab:datasets}, we list all datasets used in the experiments. Some datasets present more than one class, and in these cases the datasets were binarized to construct the tasks. To define a task, we select a dataset and one class as foreground. The pixels of all remaining classes are treated as background.

In order to assess the performance of WeaSeL in a certain setting, we employ a Leave-One-Task-Out methodology. That is, all tasks but the pair $(dataset, class)$ chosen as the Few-shot task ($\mathcal{F}$) are used in the Meta-Dataset, reserving $\mathcal{F}$ for the tuning/testing phase. This strategy serves to simultaneously hide the target task from the meta-training, while also allowing the experiments to evaluate the proposed algorithm and baselines in a myriad of scenarios. Such scenarios include: 1) target tasks with both large (e.g. JSRT~\cite{shiraishi2000development}) and small (e.g. Montgomery~\cite{jaeger2014two}) domain shifts compared to the ones in the meta-training set; 2) $\mathcal{F}$ tasks with image samples seen in other tasks used during meta-training, but with different target classes (e.g. JSRT \cite{shiraishi2000development}); and 3) $\mathcal{F}$ tasks with foreground class absent in all other tasks used during meta-training (e.g. JSRT heart segmentation).

\begin{table}[h!]
    \caption{List of datasets included in the meta-dataset.}
    \label{tab:datasets}
    \begin{adjustbox}{max width=\columnwidth}
        \begin{tabular}{@{}lccc@{}}
        \toprule
        \multicolumn{1}{c}{\textbf{Dataset}}                                     & \textbf{Image Type} & \textbf{\# of Images} & \textbf{Classes} \\ \midrule
        \textbf{JSRT~\citep{JSRTshiraishi2000development}} &
          X-rays &
          247 &
          \begin{tabular}[c]{@{}c@{}}Lungs, Clavicles\\ and Hearts\end{tabular} \\
        \textbf{Montgomery~\citep{jaeger2014two}}                                & X-rays              & 138                  & Lungs            \\
        \textbf{Shenzhen~\citep{jaeger2014two}}                                  & X-rays              & 662                  & Lungs            \\
        \textbf{NIH-labeled~\cite{NIHtang2019xlsor}}                             & X-rays              & 100                  & Lungs            \\
        \textbf{OpenIST~\cite{pi-null-mezon}} & X-rays              & 225                  & Lungs            \\
        \textbf{LIDC-IDRI-DRR~\citep{LIDColiveira20203d}}                        & CT-scans            & 835                  & Ribs             \\
        \textbf{MIAS~\citep{MIASsuckling1994mammographic}} &
          Mammograms &
          322 &
          \begin{tabular}[c]{@{}c@{}}Pectoral Muscle,\\ Breasts\end{tabular} \\
        \textbf{INbreast~\citep{moreira2012inbreast}} &
          Mammograms &
          410 &
          \begin{tabular}[c]{@{}c@{}}Pectoral Muscle,\\ Breasts\end{tabular} \\
        \textbf{Panoramic~\citep{PANORAMICabdi2015automatic}} &
          \begin{tabular}[c]{@{}c@{}}X-rays\end{tabular} &
          116 &
          Mandibles \\
        \textbf{UFBA-UESC~\citep{IVISIONsilva2018automatic}} &
          \begin{tabular}[c]{@{}c@{}}X-rays\end{tabular} &
          1500 &
          Teeths \\ \bottomrule
        \end{tabular}
    \end{adjustbox}
\end{table}

\subsection{Architecture and Hyperparameters}
\label{sec:hyperparameters}


Due to the large computational budget of second-order optimization, we propose an architecture called miniUNet\footnote{The code for WeaSeL, including the miniUNet architecture, will be available on GitHub upon the publication of this paper.}. It is a adaptation of the usual U-Net architecture with minor changes. The network is comprised of three encoder blocks, a center block, three decoder blocks and a $1\times1$ convolutional layer that works as a pixel-classification layer.
Similar to the U-Net architecture skip connections between symmetric layers are present in miniUNet, with each decoder block receiving the concatenation of the last block output and its corresponding encoder output as inputs.

\subsection{Evaluation Protocol and Metrics}
\label{sec:protocol}

We use a 5-fold cross validation protocol in the experiments, wherein datasets were divided in training and validation sets for each fold. Support sets are obtained from the training partition, while the query sets are the entire validation partition. All images and labels are resized to $128\times128$ prior to being fed to the models in order to standardize the input size and minimize the memory footprint of MAML on high-dimensional outputs. The metric within a fold is computed for all images in the query set according to the dense labels, resulting in the final values reported in Section~\ref{sec:results}, which are computed by averaging the performance across all folds. The metric used is the Jaccard score (or Intersection over Union -- IoU) of the validation images, a common metric for semantic segmentation.





\subsection{Baselines and Sparsity Modalities}
\label{sec:baselines}

We use two baselines: 1) \textit{From Scratch} training on the sparse labels; and 2) \textit{Fine-Tuning} a pretrained model from a source dataset with \textit{dense} labels. During the meta-training phase of WeaSeL, the tuning phase of \textit{Fine-Tuning}, and \textit{From Scratch} training, the sparse labels are simulated for each sample from their the dense mask.

Although being a few-shot sparse segmentation method, we do not present the Guided Nets \cite{rakelly2018conditional} as a baseline. Even to the best of our efforts, the episodic training of the original model was not able to converge to a usable model with the same medical Meta-Dataset used by WeaSeL. Thus, it did not seem fair to compare the Guided Nets to our approach.


We evaluate two types of sparse labeling, namely, \textit{points} and \textit{grid}, as well as their comparison to the performance of the models trained on the full masks. The \textit{points} labels are a selection from pixels of the image, where the annotator alternately chooses pixels from the foreground and background. In the \textit{grid} labeling the annotator receives a pre-selected group of pixels and change the class of the ones they consider positive. These pixels are disposed in a grid manner in the image. A visualization of these styles are presented in Figure~\ref{fig:sparse_modalities}.

We simulate this two types of annotations from ground truth labels. For the \textit{points} labels, given a parameter $n$, we randomly select $n$ pixels from the foreground class, and $n$ from the background. For the \textit{grid} annotation, given a parameter $s$, we choose pixels spaced, horizontally and vertically, by $s$, starting from a random distance in the range $(0-s)$ from the upper-left corner. For a consistent evaluation, a random seed for the Few-Shot $\F$ task labels is fixed for each image, ensuring that all methods use the same sparse labels for the fine-tuning process. In experiments, we vary the parameters $n$ and $s$, and analyze the impact of the sparsity of the labels.

\renewcommand{\currprop}{0.75\columnwidth}

\begin{figure}[!t]
    \centering
    \includegraphics[width=\currprop]{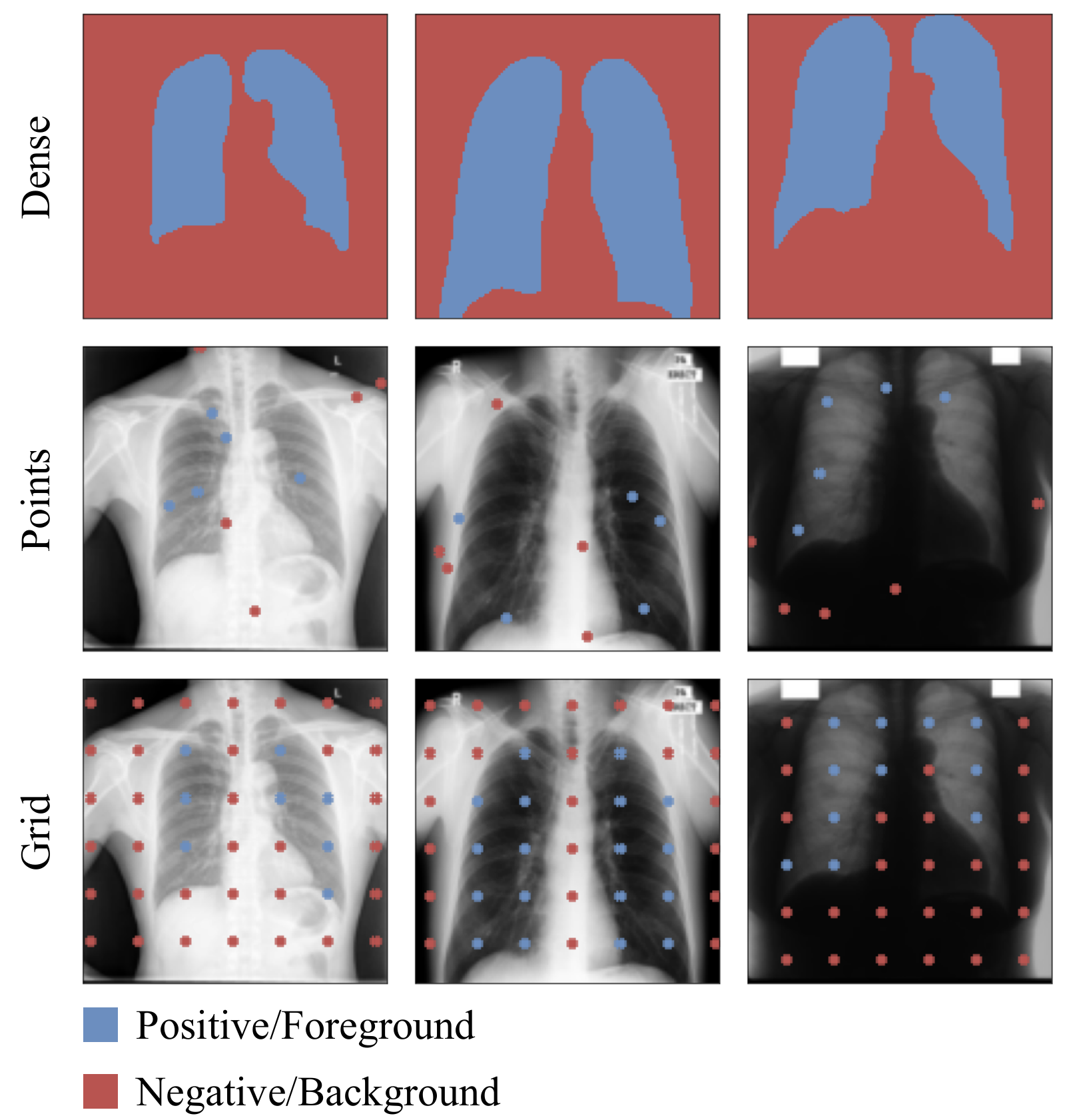}
    \caption{Example of the evaluated type of annotations. \textbf{Top Row}: the ground truth labels of all pixels. \textbf{Middle Row}: the \textit{points} annotation, $n$(5) pixels of background/foreground are labeled. \textbf{Bottom Row}: the \textit{grid} annotation, pixels spaced by $s$(20) are selected and properly labeled.}
    \label{fig:sparse_modalities}
\end{figure}

\section{Results}
\label{sec:results}


\begin{figure}[!t]
    \centering
    \centering
    \subfloat[JSRT Lungs] {
        \includegraphics[clip, trim=0.0in 0.0in 0.0in 0.0in, page=1, width=\columnwidth]{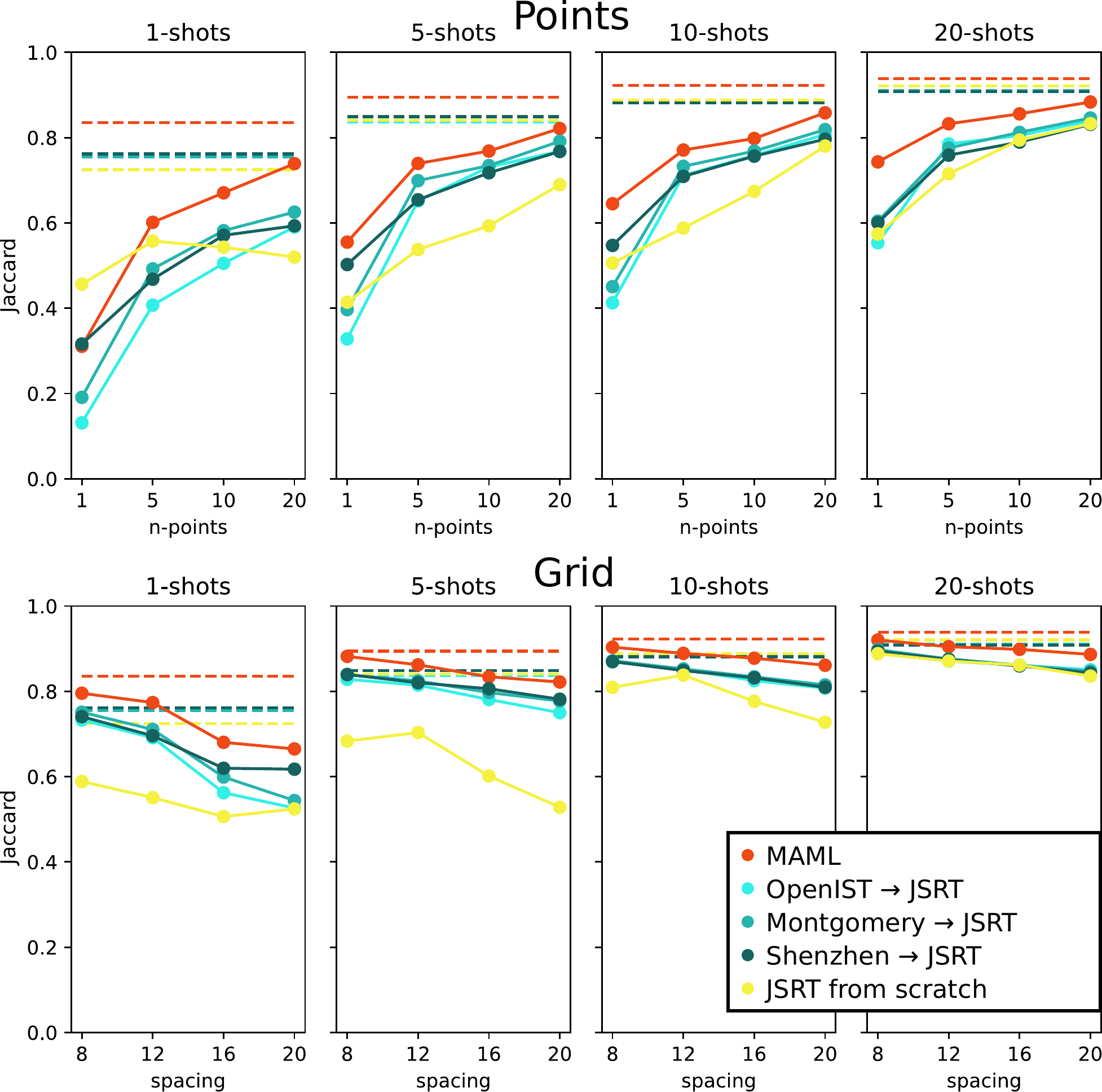}
        \label{fig:results_jsrt_lungs}
    }
    \hfill
    \subfloat[OpenIST Lungs] {
        \includegraphics[clip, trim=0.0in 0.0in 0.0in 0.0in, page=1, width=\columnwidth]{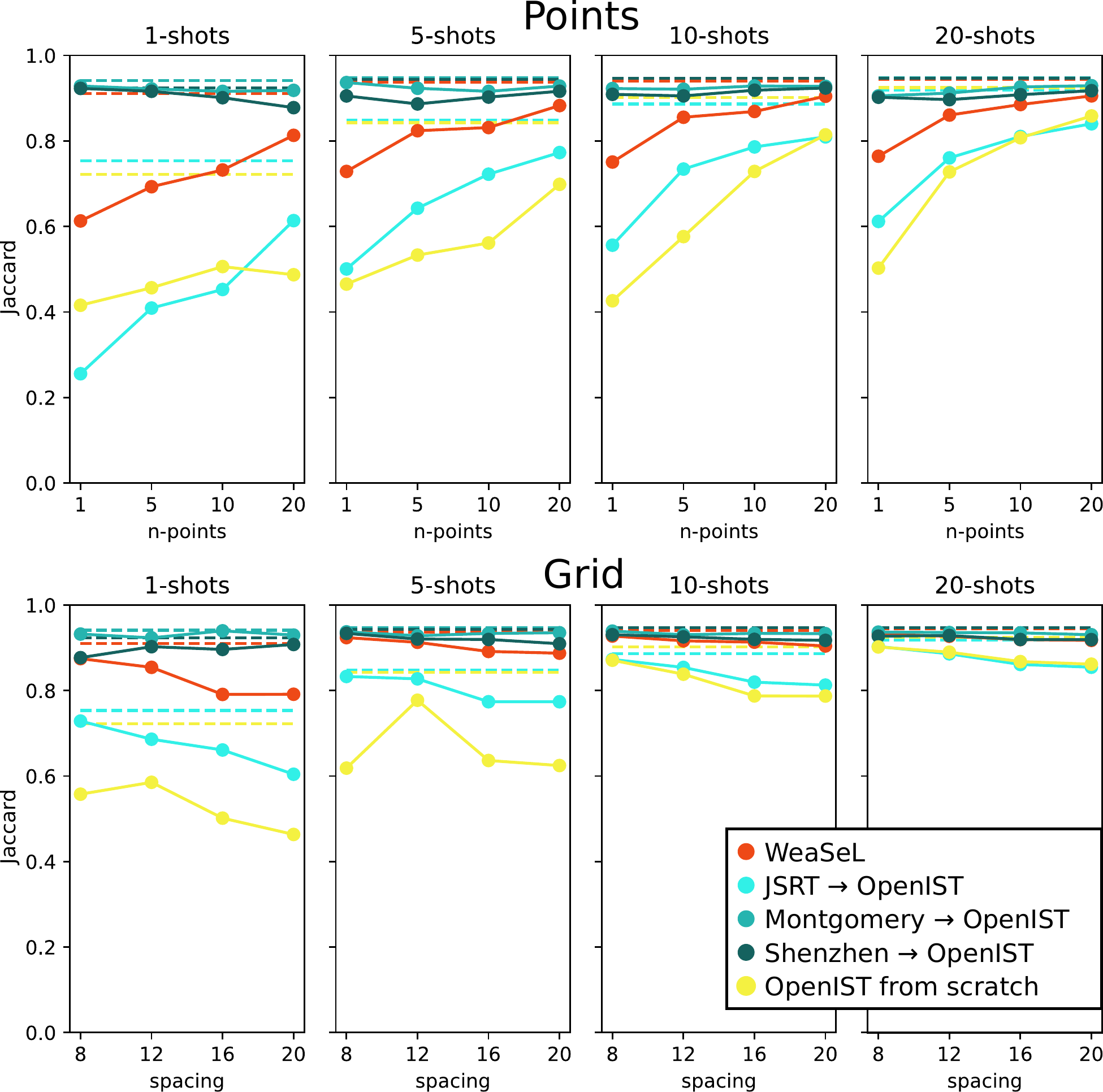}
        \label{fig:results_openist_lungs}
    }
    \caption{Jaccard results for lung segmentation in two target datasets: JSRT (a) and OpenIST (b). Solid lines indicate the performance of the methods using sparse label (\textit{Points} on the top and \textit{Grid} on the bottom), while the dashed line presents the performance of the methods trained with the dense masks.}
    \label{fig:results_lungs}
\end{figure}


\begin{figure}[!t]
    \centering
    \includegraphics[clip, trim=0.0in 0.0in 0.0in 0.0in, page=1, width=\columnwidth]{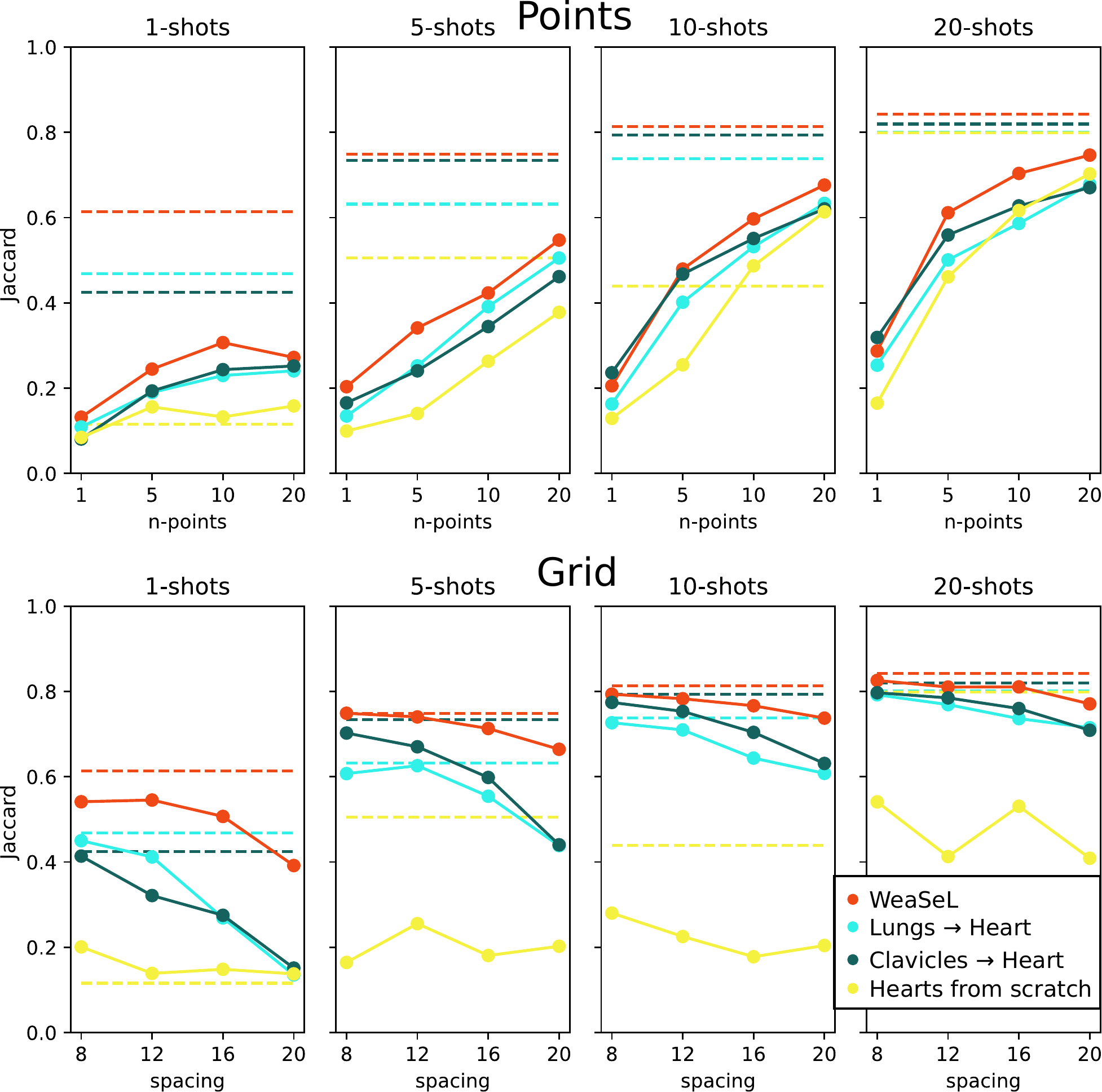}
    \caption{Jaccard results for heart segmentation in JSRT dataset.}
    \label{fig:results_jsrt_heart}
\end{figure}


We perform experiments to access the performance of WeaSeL in all CXR datasets. That is, we evaluate the Few-shot task segmentation of lungs in all five datasets, and hearts and clavicles segmentation in the JSRT dataset. 
For brevity, we include only a subset of the results.

In Figure~\ref{fig:results_lungs} we see the results of experiments in the lungs class. The results of Montgomery and Shenzhen datasets are similar to the OpenIST, thus we only present the later. WeaSeL shows better metrics than the \textit{From Scratch} baseline in all cases. As expected, increasing the amount of annotated pixels (e.g. by increasing $n$, or decreasing the grid spacing $s$), as well as having more training samples (larger $k$-shots), have a direct impact in IoU. For JSRT (Figure~\ref{fig:results_jsrt_lungs}), WeaSeL yields better performance than \textit{Fine-Tuning} from all datasets, in contrast with OpenIST (Figure~\ref{fig:results_openist_lungs}) where the source dataset was decisive in the performance of \textit{Fine-Tuning}. The networks pretrained on JSRT achieved performances lower than WeaSeL, while the pretraining from Shenzhen or Montgomery yielded better results than our approach. This discrepancy can be explained by the different domains, as the JSRT dataset is the most singular within the five CXR datasets -- that is, the domain shift between JSRT and the other datasets is larger than the shift between the other four. It should be noticed that real-world scenarios may not allow for a fair evaluation of the domain shift between the source and target datasets, making WeaSeL the safer choice in this case. 

In Figure~\ref{fig:results_jsrt_heart} we see the results for the heart class in JSRT. This, and the clavicles class, are only present in the JSRT dataset. 
Since these classes are only present in JSRT, for the \textit{Fine-Tuning} baselines we choose the pairs of JSRT dataset and one of the remaining classes as source tasks (e.g., for hearts we use the pairs $(JSRT, lungs)$, and $(JSRT, clavicles)$).
Again, the \textit{From Scratch} baseline is the worst performer, with the WeaSeL being the superior in most scenarios. The clavicles, display a similar tendency, although their IoU score, in majority of the cases, is lower than the heart class scores.

\begin{figure}[!t]
    \centering
    \includegraphics[clip, trim=0.0in 0.0in 0.0in 0.0in, page=1, width=\columnwidth]{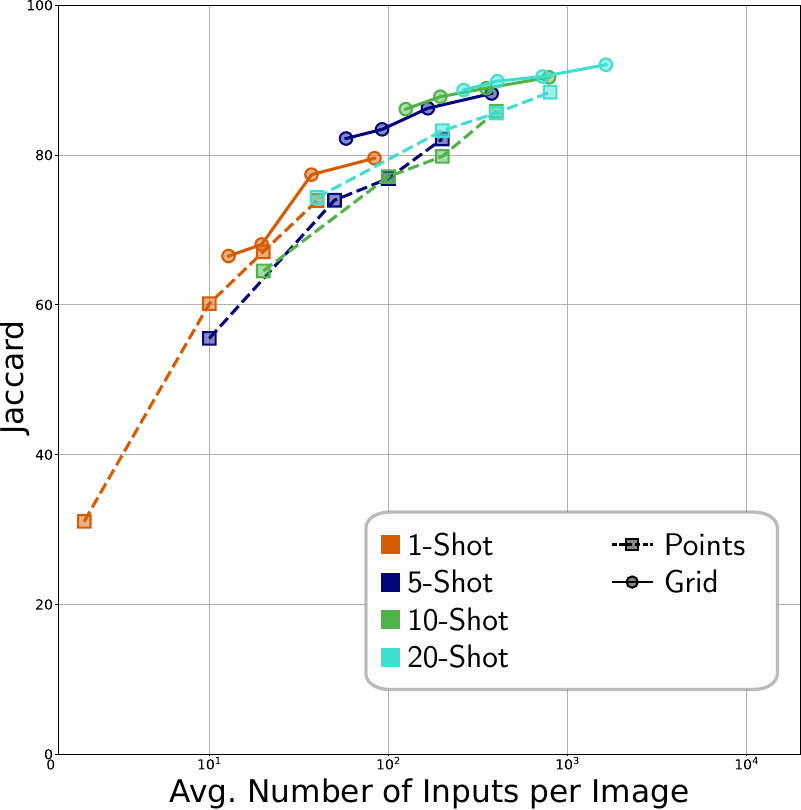}
    \caption{Jaccard results of the WeaSeL method in the JSRT Lungs task by the average number of inputs.}
    \label{fig:inputs_jsrt_lungs}
\end{figure}


As seen in both experiments, the performance of the \textit{grid} annotation is higher than the \textit{points} annotation, in virtually all cases. In order to assess the efficiency of this annotations we constructed the graph in Figure~\ref{fig:inputs_jsrt_lungs}. The \textit{x} axis represents the average number of inputs per image, and was computed considering each positive labeled pixel in an image as an user input, and then average across all images. In the figure, we confirm the overall better performance of the \textit{grid} annotation, and a better efficiency as the score of the method with the same number of inputs is higher using this annotation over the \textit{point} annotation. One explanation is that even in the larger spacing scenario($s=20$) the number of annotated pixels is greater than the the $2n$ pixels in any \textit{points} labeling case. Thus, even with a likely class imbalance in the \textit{grid} scenario --- the foreground objects is usually smaller than the background, and with a fixed grid will probably have fewer labeled pixels ---, the simple presence of more data increases the metric score for this type of annotation.

Sparse annotation show competitive results with dense labels, given the much smaller labeled data.
The case with most labeled data is the \textit{grid} annotation with $s=20$, that have approximately $~2\%$ of annotated pixels. In some cases, the methods using sparse labels are equivalent to using dense labels ground truths, specially with the \textit{grid} annotation. In the lungs segmentation in the OpenIST (and Montgomery/Shenzhen) dataset (Figure~\ref{fig:results_openist_lungs}), the \textit{Fine-tuning} baselines from similar datasets seem indifferent to sparse or dense labels. As aforementioned, the small domain shifts between these datasets and the pretraining with dense labels explain this results.

\section{Qualitative Analysis}
\label{sec:qualitative}

\begin{figure*}[h!]
    \centering
    \includegraphics[width=0.85\textwidth]{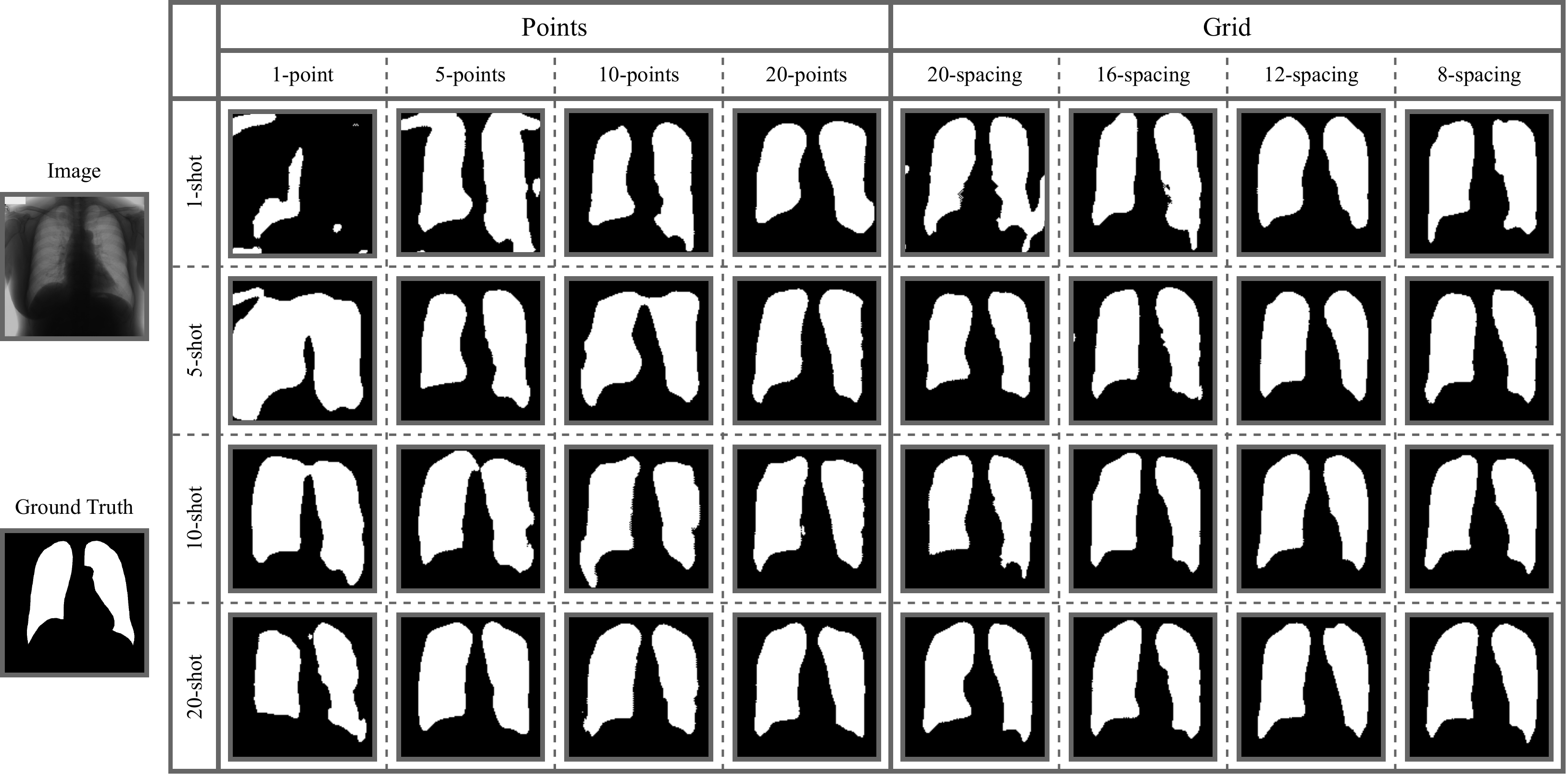}
    \caption{Visual segmentation examples for the JSRT Lungs task.}
    \label{fig:jsrt_lungs_visual}
\end{figure*}

\begin{figure*}[h!]
    \centering
    \includegraphics[width=0.85\textwidth]{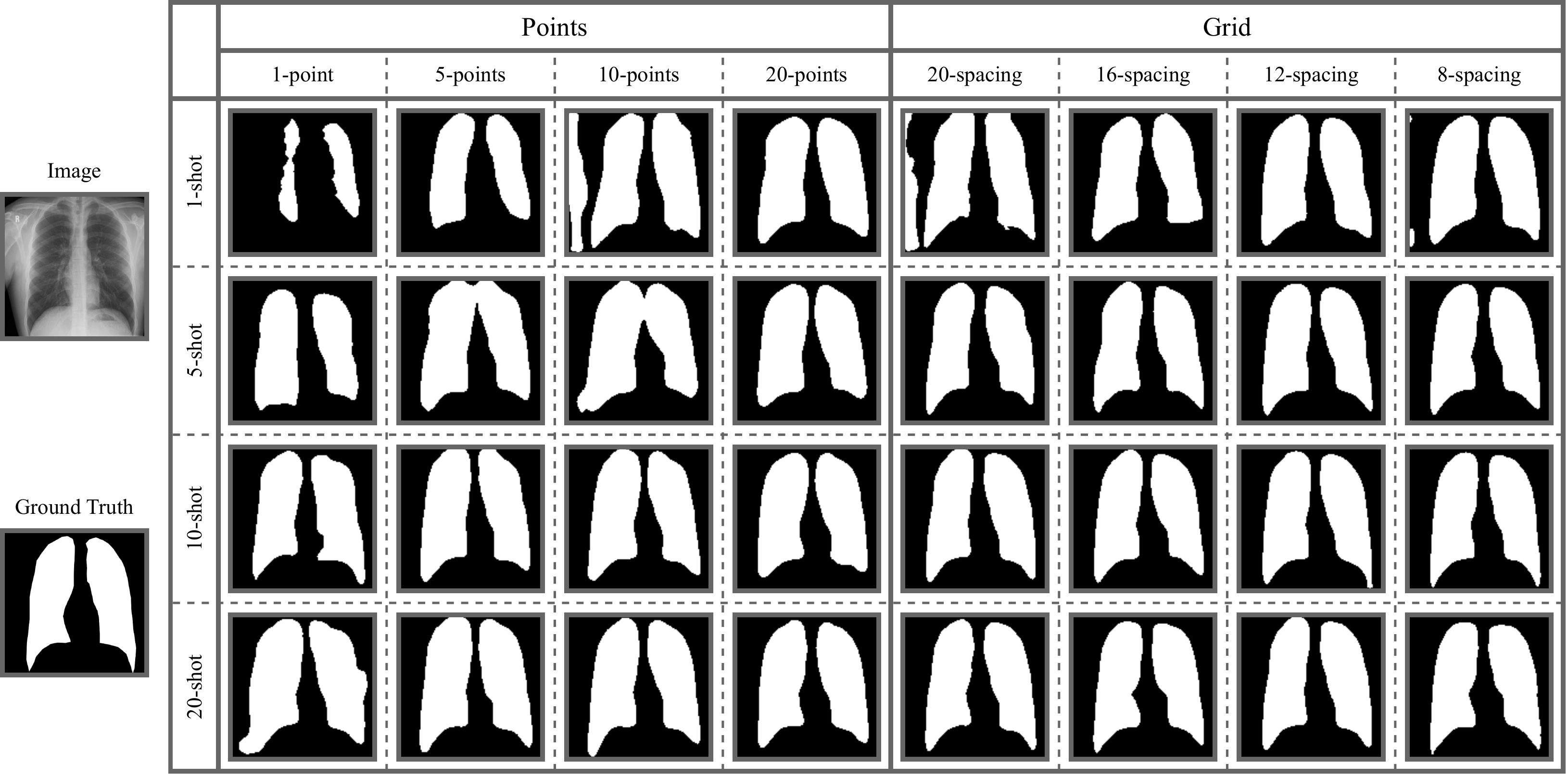}
    \caption{Visual segmentation examples for the OpenIST Lungs task.}
    \label{fig:openist_lungs_visual}
\end{figure*}

\begin{figure*}[h!]
    \centering
    \includegraphics[width=0.85\textwidth]{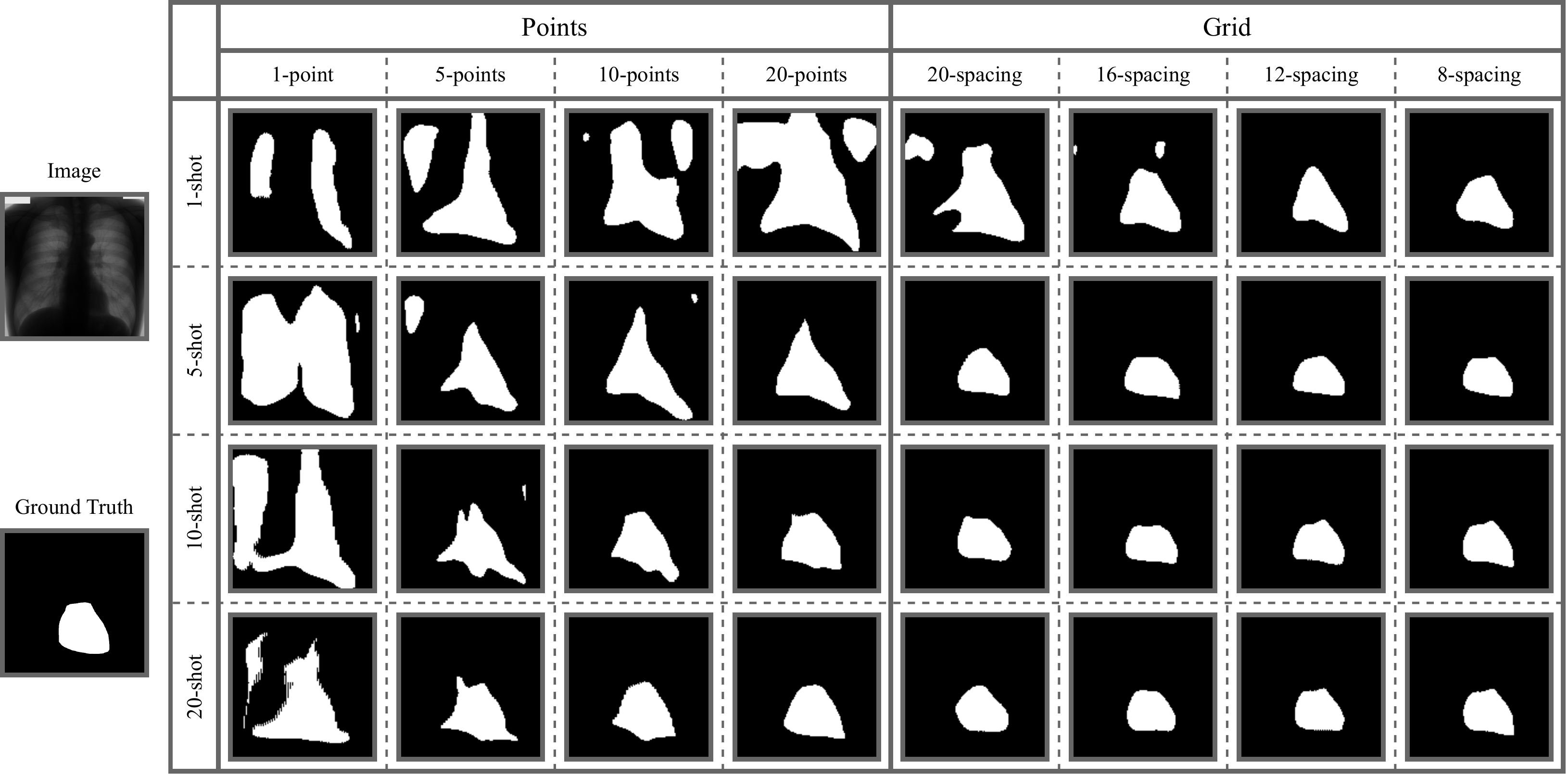}
    \caption{Visual segmentation examples for the JSRT Heart task.}
    \label{fig:jsrt_heart_visual}
\end{figure*}

In Figures~\ref{fig:jsrt_lungs_visual}-\ref{fig:jsrt_heart_visual} we show visual segmentation examples of WeaSeL on three tasks: JSRT Lungs, OpenIST Lungs and JSRT Heart, respectively. In each example, lines represent the number of shots in the experiment, while columns vary the sparsity parameters: $n$ for points (first four columns) and $s$ for grids (last four columns). These sparsity scenarios were used to simulate the sparse labels of the support set on the target few-shot task.

As one can observe in these visual results, increasing the annotation density -- that is, increasing the number of points $n$ or using a smaller spacing $s$ -- yields a larger improvement in the predictions than increasing the number of labeled images; that is, the number of shots. Even the 1-shot scenario achieved acceptable results for the lung tasks (Figures~\ref{fig:jsrt_lungs_visual},~\ref{fig:openist_lungs_visual}) when given a sufficient number of labeled pixels, e.g. at least 20 labeled pixels for each class.
For heart segmentation (Figure~\ref{fig:jsrt_heart_visual}), we observe a great difference between the points and grid annotations. Since this is a unique task -- i.e. the only task in the meta-dataset with \textit{heart} as a target class, which is only present in the JSRT dataset -- the model requires more support samples/labels to adapt to this task. This requirement is not fully achieved by the points annotation with one and five shots, but is obtained by less sparse grid annotations in all observed number of shots. Again, the total number of labeled pixels in the grid annotation seems to compensate the lack of images for training.

It is clear that having a larger number of annotated images impacts the final results, especially for harder, or more distinct, datasets or tasks as is the case of JSRT. Consequently, a middle ground in relation to the number of images and the sparsity of annotations seems to be a good compromise to achieve reasonable results with limited human intervention. That is, annotating five or more images with at least five pixels labeled per class can lead to good results with a very low labeling burden. However, if the target task is truly few-shot, increasing the number of samples in the support set may be costly or even impossible. In this case, one can instead increase the per sample label density.

\newcommand\cincludegraphics[2][]{\raisebox{-0.1\height}{\includegraphics[#1]{#2}}}
\newcommand\exImages{0.15}

\section{Conclusion}
\label{sec:conclusion}

We evaluated WeaSeL, our proposed adaptation of MAML, in multiple few-shot segmentation tasks with sparse labels. WeaSeL showed to be a viable solution to most tasks, and a suitable option when there are very discrepant target and source tasks. Since in real-world scenarios is difficult to access the domain shift between the datasets, WeaSeL seems to be the safer choice in these cases. The method have some limitations, like the cost of computing second-derivatives in the outer loop. This limited the size of training images to $128\times128$, which can hinder task with small target objects (e.g. clavicles). 

Future works include the use of approximations to the second derivative and how this affects the performance, as well as analyzing different types of sparse labels annotations also in 3D volumes. Additionally, we intend to adapt other first- and second-order Meta-Learning methods for sparsely labeled semantic segmentation.




\bibliographystyle{abbrvnat}
\bibliography{references}

\end{document}


\title{Learning to Segment Medical Images from Few-Shot Sparse Labels: Suplementary Material}

\author{Pedro H. T. Gama \qquad Hugo Oliveira \qquad Jefersson A. dos Santos
}
\date{2021}

\maketitle

\pagestyle{myheadings}

\section{Additional results}

\begin{table}[h!]
\caption{Jaccard Score for Montgomery Lungs segmentation.}
\label{tab:montgomery}
\centering
\adjustbox{pagecenter, width=4.5in}{
\begin{tabular}{@{}ccccccccccccccccccccc@{}}
\toprule
\multicolumn{1}{l}{\textbf{}} & \multicolumn{4}{c}{\textbf{From Scratch}} & \multicolumn{4}{c}{\textbf{From JSRT}} & \multicolumn{4}{c}{\textbf{From OpenIST}} & \multicolumn{4}{c}{\textbf{From Shenzhen}} & \multicolumn{4}{c}{\textbf{WeaSeL}} \\ \midrule
\multicolumn{1}{c|}{} & \multicolumn{4}{c|}{\textbf{Points}} & \multicolumn{4}{c|}{\textbf{Points}} & \multicolumn{4}{c|}{\textbf{Points}} & \multicolumn{4}{c|}{\textbf{Points}} & \multicolumn{4}{c}{\textbf{Points}} \\ \midrule
\multicolumn{1}{c|}{\textbf{Shot}} & \textbf{1} & \textbf{5} & \textbf{10} & \multicolumn{1}{c|}{\textbf{20}} & \textbf{1} & \textbf{5} & \textbf{10} & \multicolumn{1}{c|}{\textbf{20}} & \textbf{1} & \textbf{5} & \textbf{10} & \multicolumn{1}{c|}{\textbf{20}} & \textbf{1} & \textbf{5} & \textbf{10} & \multicolumn{1}{c|}{\textbf{20}} & \textbf{1} & \textbf{5} & \textbf{10} & \textbf{20} \\
\multicolumn{1}{c|}{\textbf{1}} & \multicolumn{1}{l}{28.64} & \multicolumn{1}{l}{31.26} & \multicolumn{1}{l}{35.99} & \multicolumn{1}{l|}{39.91} & 15.71 & 29.78 & 43.19 & \multicolumn{1}{c|}{49.71} & 88.42 & 84.54 & 84.82 & \multicolumn{1}{c|}{85.89} & 82.26 & 78.84 & 79.49 & \multicolumn{1}{c|}{84.08} & 43.23 & 63.28 & 77.49 & 78.40 \\
\multicolumn{1}{c|}{\textbf{5}} & \multicolumn{1}{l}{28.93} & \multicolumn{1}{l}{33.34} & \multicolumn{1}{l}{46.75} & \multicolumn{1}{l|}{60.65} & 28.92 & 50.63 & 63.43 & \multicolumn{1}{c|}{71.06} & 83.96 & 86.24 & 88.79 & \multicolumn{1}{c|}{89.16} & 80.31 & 84.09 & 86.79 & \multicolumn{1}{c|}{89.14} & 59.67 & 80.21 & 81.55 & 85.26 \\
\multicolumn{1}{c|}{\textbf{10}} & \multicolumn{1}{l}{30.52} & \multicolumn{1}{l}{48.87} & \multicolumn{1}{l}{66.53} & \multicolumn{1}{l|}{76.56} & 38.16 & 64.46 & 73.88 & \multicolumn{1}{c|}{78.73} & 87.54 & 88.69 & 89.51 & \multicolumn{1}{c|}{90.15} & 84.38 & 86.54 & 89.19 & \multicolumn{1}{c|}{90.97} & 69.64 & 78.92 & 84.54 & 87.05 \\
\multicolumn{1}{c|}{\textbf{20}} & \multicolumn{1}{l}{39.15} & \multicolumn{1}{l}{63.05} & \multicolumn{1}{l}{77.65} & \multicolumn{1}{l|}{82.16} & 52.73 & 73.82 & 79.81 & \multicolumn{1}{c|}{83.29} & 89.48 & 90.15 & 90.02 & \multicolumn{1}{c|}{91.33} & 86.07 & 89.46 & 89.07 & \multicolumn{1}{c|}{91.20} & 74.17 & 85.01 & 87.80 & 89.02 \\ \midrule
\multicolumn{1}{c|}{\textbf{}} & \multicolumn{4}{c|}{\textbf{Grid}} & \multicolumn{4}{c|}{\textbf{Grid}} & \multicolumn{4}{c|}{\textbf{Grid}} & \multicolumn{4}{c|}{\textbf{Grid}} & \multicolumn{4}{c}{\textbf{Grid}} \\ \midrule
\multicolumn{1}{c|}{\textbf{Shot}} & \textbf{8} & \textbf{12} & \textbf{16} & \multicolumn{1}{c|}{\textbf{20}} & \textbf{8} & \textbf{12} & \textbf{16} & \multicolumn{1}{c|}{\textbf{20}} & \textbf{8} & \textbf{12} & \textbf{16} & \multicolumn{1}{c|}{\textbf{20}} & \textbf{8} & \textbf{12} & \textbf{16} & \multicolumn{1}{c|}{\textbf{20}} & \textbf{8} & \textbf{12} & \textbf{16} & \textbf{20} \\
\multicolumn{1}{c|}{\textbf{1}} & 46.92 & 36.34 & 37.81 & \multicolumn{1}{c|}{40.40} & 62.90 & 59.17 & 56.89 & \multicolumn{1}{c|}{49.50} & 91.50 & 90.61 & 92.08 & \multicolumn{1}{c|}{92.45} & 92.89 & 91.07 & 91.23 & \multicolumn{1}{c|}{92.54} & 87.81 & 83.72 & 82.68 & 78.03 \\
\multicolumn{1}{c|}{\textbf{5}} & 50.22 & 58.43 & 54.46 & \multicolumn{1}{c|}{48.80} & 82.28 & 77.55 & 75.42 & \multicolumn{1}{c|}{71.00} & 93.46 & 92.06 & 91.64 & \multicolumn{1}{c|}{91.91} & 93.58 & 91.18 & 91.46 & \multicolumn{1}{c|}{92.32} & 91.74 & 89.86 & 88.43 & 86.12 \\
\multicolumn{1}{c|}{\textbf{10}} & 70.54 & 79.54 & 77.58 & \multicolumn{1}{c|}{59.96} & 85.30 & 83.05 & 80.90 & \multicolumn{1}{c|}{76.87} & 92.90 & 92.54 & 91.63 & \multicolumn{1}{c|}{92.16} & 92.94 & 92.24 & 92.09 & \multicolumn{1}{c|}{92.62} & 92.05 & 91.14 & 89.93 & 87.71 \\
\multicolumn{1}{c|}{\textbf{20}} & 88.70 & 87.11 & 85.47 & \multicolumn{1}{c|}{83.13} & 88.99 & 87.27 & 85.41 & \multicolumn{1}{c|}{83.62} & 93.40 & 93.24 & 92.84 & \multicolumn{1}{c|}{92.78} & 93.23 & 92.59 & 92.12 & \multicolumn{1}{c|}{92.04} & 92.89 & 92.14 & 91.54 & 90.36
\end{tabular}
}
\end{table}

\begin{table}[h!]
\caption{Jaccard Score for Shenzhen Lungs segmentation.}
\label{tab:shenzhen}
\centering
\adjustbox{pagecenter, width=4.5in}{
\begin{tabular}{@{}ccccccccccccccccccccc@{}}
\toprule
\multicolumn{1}{l}{\textbf{}} & \multicolumn{4}{c}{\textbf{From Scratch}} & \multicolumn{4}{c}{\textbf{From JSRT}} & \multicolumn{4}{c}{\textbf{From OpenIST}} & \multicolumn{4}{c}{\textbf{From Montgomery}} & \multicolumn{4}{c}{\textbf{WeaSeL}} \\ \midrule
\multicolumn{1}{c|}{} & \multicolumn{4}{c|}{\textbf{Points}} & \multicolumn{4}{c|}{\textbf{Points}} & \multicolumn{4}{c|}{\textbf{Points}} & \multicolumn{4}{c|}{\textbf{Points}} & \multicolumn{4}{c}{\textbf{Points}} \\ \midrule
\multicolumn{1}{c|}{\textbf{Shot}} & \textbf{1} & \textbf{5} & \textbf{10} & \multicolumn{1}{c|}{\textbf{20}} & \textbf{1} & \textbf{5} & \textbf{10} & \multicolumn{1}{c|}{\textbf{20}} & \textbf{1} & \textbf{5} & \textbf{10} & \multicolumn{1}{c|}{\textbf{20}} & \textbf{1} & \textbf{5} & \textbf{10} & \multicolumn{1}{c|}{\textbf{20}} & \textbf{1} & \textbf{5} & \textbf{10} & \textbf{20} \\
\multicolumn{1}{c|}{\textbf{1}} & 32.26 & 35.75 & 38.07 & \multicolumn{1}{c|}{43.05} & 27.11 & 38.09 & 47.46 & \multicolumn{1}{c|}{58.88} & 83.42 & 81.27 & 86.57 & \multicolumn{1}{c|}{86.03} & 81.99 & 79.33 & 84.82 & \multicolumn{1}{c|}{84.37} & 53.40 & 59.86 & 71.58 & 74.80 \\
\multicolumn{1}{c|}{\textbf{5}} & 36.66 & 47.53 & 54.35 & \multicolumn{1}{c|}{58.12} & 51.78 & 65.35 & 67.90 & \multicolumn{1}{c|}{69.09} & 87.35 & 87.31 & 88.46 & \multicolumn{1}{c|}{85.61} & 85.90 & 86.96 & 85.42 & \multicolumn{1}{c|}{83.70} & 68.91 & 82.97 & 81.78 & 81.26 \\
\multicolumn{1}{c|}{\textbf{10}} & 40.57 & 55.62 & 65.17 & \multicolumn{1}{c|}{71.85} & 58.36 & 69.33 & 72.92 & \multicolumn{1}{c|}{75.01} & 87.74 & 85.51 & 87.51 & \multicolumn{1}{c|}{85.48} & 86.24 & 85.29 & 86.26 & \multicolumn{1}{c|}{84.14} & 69.82 & 80.57 & 82.58 & 83.00 \\
\multicolumn{1}{c|}{\textbf{20}} & 48.14 & 65.88 & 73.50 & \multicolumn{1}{c|}{77.62} & 61.15 & 74.01 & 77.04 & \multicolumn{1}{c|}{78.09} & 86.31 & 84.70 & 85.74 & \multicolumn{1}{c|}{84.94} & 83.03 & 84.82 & 84.84 & \multicolumn{1}{c|}{83.10} & 71.44 & 80.71 & 82.39 & 82.78 \\ \midrule
\multicolumn{1}{c|}{\textbf{}} & \multicolumn{4}{c|}{\textbf{Grid}} & \multicolumn{4}{c|}{\textbf{Grid}} & \multicolumn{4}{c|}{\textbf{Grid}} & \multicolumn{4}{c|}{\textbf{Grid}} & \multicolumn{4}{c}{\textbf{Grid}} \\ \midrule
\multicolumn{1}{c|}{\textbf{Shot}} & \textbf{8} & \textbf{12} & \textbf{16} & \multicolumn{1}{c|}{\textbf{20}} & \textbf{8} & \textbf{12} & \textbf{16} & \multicolumn{1}{c|}{\textbf{20}} & \textbf{8} & \textbf{12} & \textbf{16} & \multicolumn{1}{c|}{\textbf{20}} & \textbf{8} & \textbf{12} & \textbf{16} & \multicolumn{1}{c|}{\textbf{20}} & \textbf{8} & \textbf{12} & \textbf{16} & \textbf{20} \\
\multicolumn{1}{c|}{\textbf{1}} & 44.67 & 54.18 & 40.81 & \multicolumn{1}{c|}{40.35} & 69.78 & 63.92 & 59.97 & \multicolumn{1}{c|}{52.63} & 84.05 & 81.50 & 83.45 & \multicolumn{1}{c|}{82.83} & 81.79 & 80.50 & 81.07 & \multicolumn{1}{c|}{81.65} & 79.90 & 76.85 & 75.89 & 78.63 \\
\multicolumn{1}{c|}{\textbf{5}} & 52.53 & 50.58 & 52.81 & \multicolumn{1}{c|}{49.43} & 75.19 & 71.59 & 71.60 & \multicolumn{1}{c|}{71.26} & 87.24 & 85.85 & 87.21 & \multicolumn{1}{c|}{85.73} & 86.40 & 84.06 & 85.74 & \multicolumn{1}{c|}{86.14} & 82.88 & 83.59 & 83.18 & 79.72 \\
\multicolumn{1}{c|}{\textbf{10}} & 73.60 & 76.38 & 69.71 & \multicolumn{1}{c|}{55.20} & 80.92 & 78.97 & 78.04 & \multicolumn{1}{c|}{75.16} & 86.99 & 86.71 & 87.01 & \multicolumn{1}{c|}{83.39} & 86.12 & 84.64 & 85.63 & \multicolumn{1}{c|}{82.96} & 85.53 & 84.09 & 82.85 & 80.97 \\
\multicolumn{1}{c|}{\textbf{20}} & 81.76 & 81.77 & 80.97 & \multicolumn{1}{c|}{76.54} & 83.56 & 82.37 & 81.43 & \multicolumn{1}{c|}{79.00} & 86.86 & 85.94 & 86.40 & \multicolumn{1}{c|}{84.55} & 85.91 & 85.26 & 84.76 & \multicolumn{1}{c|}{82.88} & 86.93 & 86.33 & 85.73 & 84.11
\end{tabular}
}
\end{table}

\begin{figure}[!h]
    \centering
    \centering
    \subfloat[Montgomery Lungs] {
        \includegraphics[clip, trim=0.0in 0.0in 0.0in 0.0in, page=1, width=0.7\textwidth]{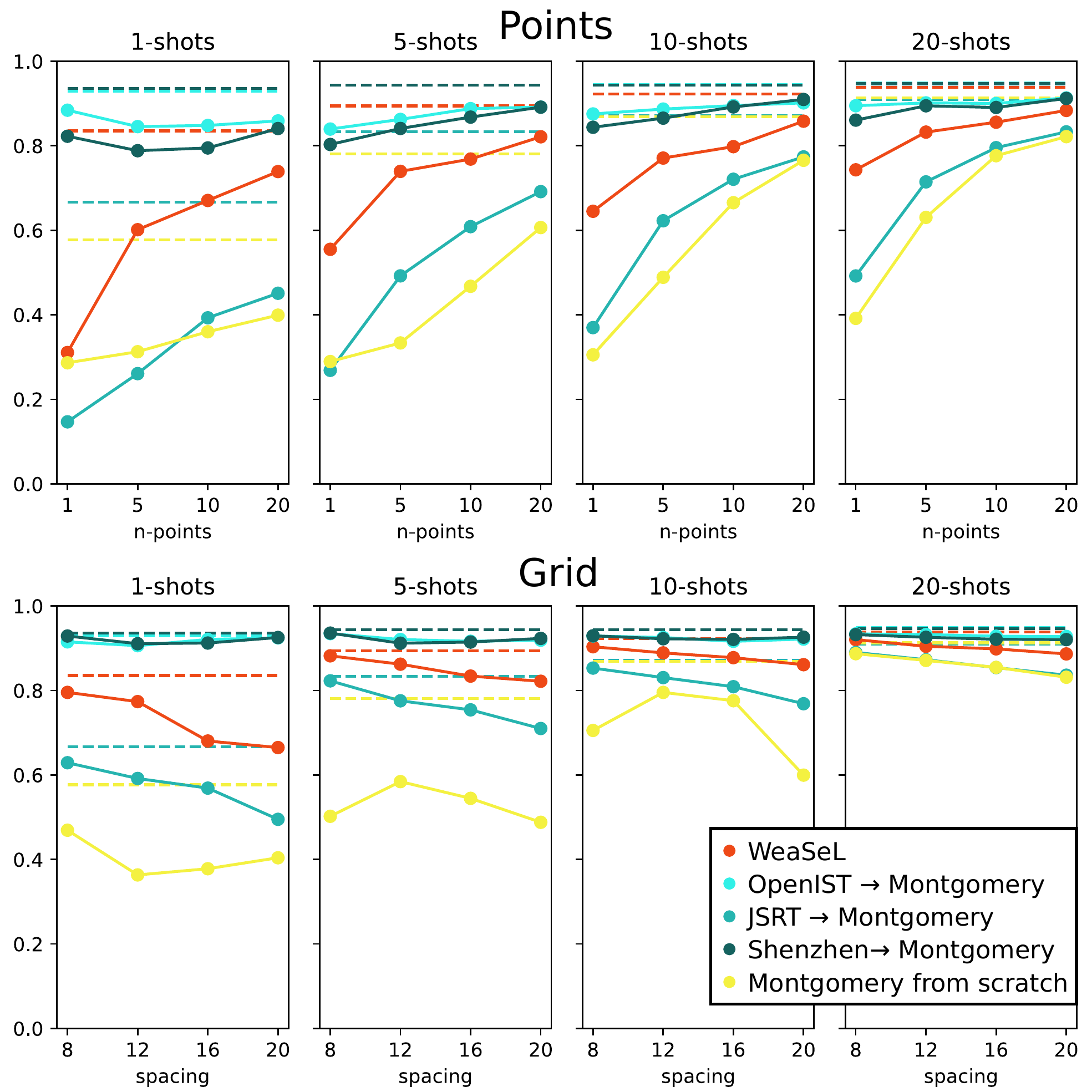}
        \label{fig:results_montgomery_lungs}
    }
    \hfill
    \subfloat[Shenzhen Lungs] {
        \includegraphics[clip, trim=0.0in 0.0in 0.0in 0.0in, page=1, width=0.7\textwidth]{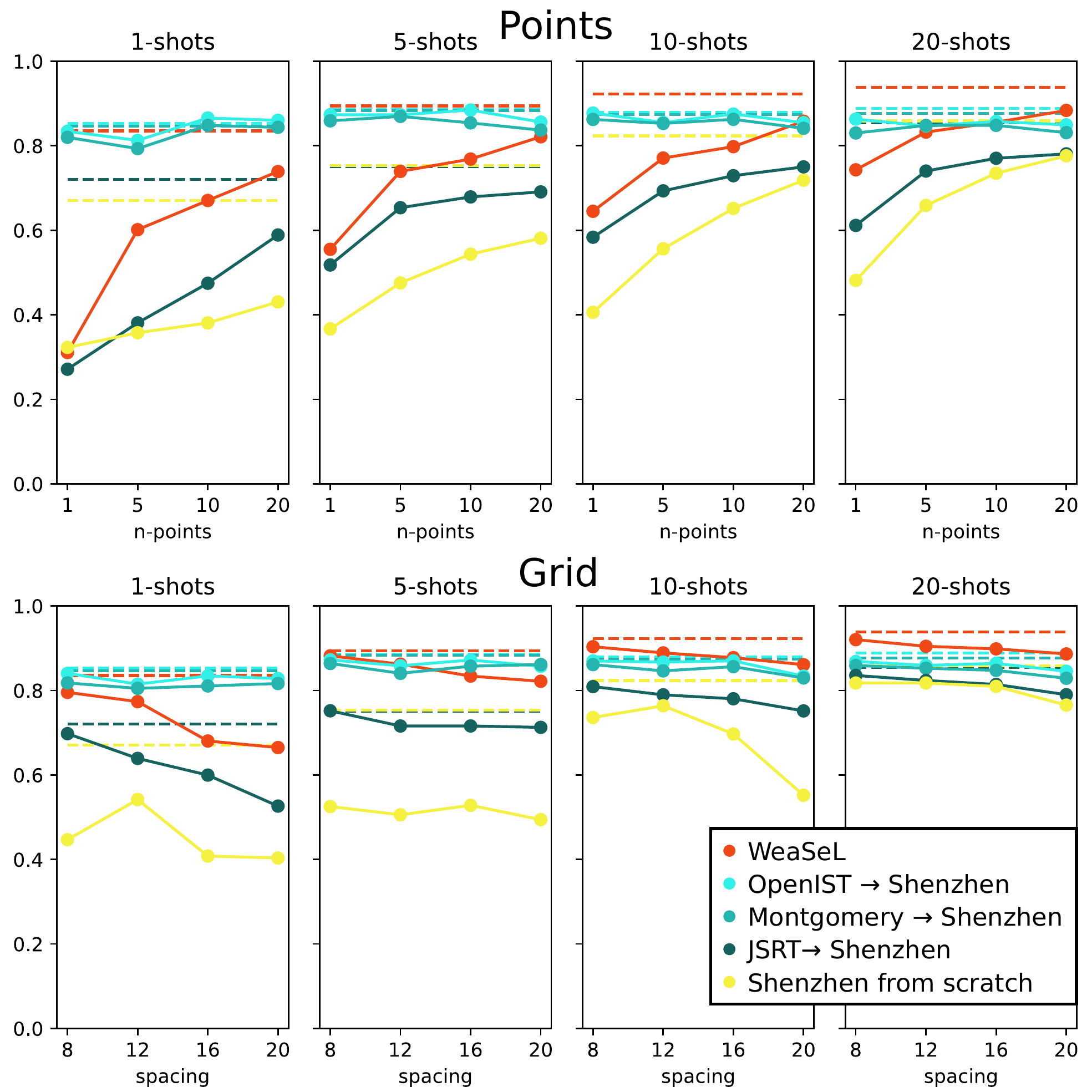}
        \label{fig:results_shenzhen_lungs}
    }
    \caption{Jaccard results for lung segmentation in two target datasets: Montgomery (a) and Shenzhen (b). Solid lines indicate the performance of the sparse labeling (\textit{Points} on the top and \textit{Grid} on the bottom), while the dashed line presents the performance of dense labeling for the target using the same methods and baselines.}
    \label{fig:sup_results_lungs}
\end{figure}

As mentioned in the paper, the results for Montgomery and Shenzhen lungs segmentation task, had results similar to the OpenIST dataset, as can be seen in tables~\ref{tab:montgomery} and~\ref{tab:shenzhen}, and in Figure~\ref{fig:sup_results_lungs}. So the same conclusions from the paper holds for these datasets: \textit{From Scratch} being the worst baseline, \textit{Fine Tuning} from similar domains providing the best results, while WeaSeL reaching comparable results to the Fine-Tuning baselines with more data samples.

\begin{table}[h!]
\caption{Jaccard Score for JSRT Clavicles segmentation.}
\label{tab:jsrt_clavicles}
\centering
\adjustbox{pagecenter, width=4.5in}{
\begin{tabular}{@{}ccccccccccccccccc@{}}
\toprule
\multicolumn{1}{l}{\textbf{}} & \multicolumn{4}{c}{\textbf{From Scratch}} & \multicolumn{4}{c}{\textbf{From JSRT Hearts}} & \multicolumn{4}{c}{\textbf{From JSRT Lungs}} & \multicolumn{4}{c}{\textbf{WeaSeL}} \\ \midrule
\multicolumn{1}{c|}{} & \multicolumn{4}{c|}{\textbf{Points}} & \multicolumn{4}{c|}{\textbf{Points}} & \multicolumn{4}{c|}{\textbf{Points}} & \multicolumn{4}{c}{\textbf{Points}} \\ \midrule
\multicolumn{1}{c|}{\textbf{Shot}} & \textbf{1} & \textbf{5} & \textbf{10} & \multicolumn{1}{c|}{\textbf{20}} & \textbf{1} & \textbf{5} & \textbf{10} & \multicolumn{1}{c|}{\textbf{20}} & \textbf{1} & \textbf{5} & \textbf{10} & \multicolumn{1}{c|}{\textbf{20}} & \textbf{1} & \textbf{5} & \textbf{10} & \textbf{20} \\
\multicolumn{1}{c|}{\textbf{1}} & 4.36 & 4.37 & 4.90 & \multicolumn{1}{c|}{5.30} & 5.32 & 6.85 & 7.73 & \multicolumn{1}{c|}{14.55} & 7.33 & 7.31 & 7.85 & \multicolumn{1}{c|}{9.13} & 7.52 & 7.65 & 10.67 & 12.04 \\
\multicolumn{1}{c|}{\textbf{5}} & 4.23 & 4.64 & 5.40 & \multicolumn{1}{c|}{13.32} & 8.98 & 16.20 & 15.41 & \multicolumn{1}{c|}{23.60} & 7.65 & 10.50 & 11.58 & \multicolumn{1}{c|}{17.38} & 7.44 & 10.77 & 15.23 & 22.04 \\
\multicolumn{1}{c|}{\textbf{10}} & 4.53 & 7.50 & 14.21 & \multicolumn{1}{c|}{22.80} & 9.49 & 18.54 & 22.66 & \multicolumn{1}{c|}{28.73} & 7.44 & 11.21 & 13.81 & \multicolumn{1}{c|}{21.38} & 8.33 & 17.38 & 24.62 & 29.63 \\
\multicolumn{1}{c|}{\textbf{20}} & 5.83 & 14.75 & 23.94 & \multicolumn{1}{c|}{31.48} & 10.63 & 21.17 & 26.06 & \multicolumn{1}{c|}{31.62} & 9.17 & 13.78 & 17.12 & \multicolumn{1}{c|}{26.95} & 11.55 & 23.58 & 33.27 & 39.58 \\ \midrule
\multicolumn{1}{c|}{\textbf{}} & \multicolumn{4}{c|}{\textbf{Grid}} & \multicolumn{4}{c|}{\textbf{Grid}} & \multicolumn{4}{c|}{\textbf{Grid}} & \multicolumn{4}{c}{\textbf{Grid}} \\ \midrule
\multicolumn{1}{c|}{\textbf{Shot}} & \textbf{8} & \textbf{12} & \textbf{16} & \multicolumn{1}{c|}{\textbf{20}} & \textbf{8} & \textbf{12} & \textbf{16} & \multicolumn{1}{c|}{\textbf{20}} & \textbf{8} & \textbf{12} & \textbf{16} & \multicolumn{1}{c|}{\textbf{20}} & \textbf{8} & \textbf{12} & \textbf{16} & \textbf{20} \\
\multicolumn{1}{c|}{\textbf{1}} & 3.05 & 3.27 & 2.87 & \multicolumn{1}{c|}{2.89} & 3.99 & .80 & 9.89 & \multicolumn{1}{c|}{9.33} & 19.43 & 20.47 & 14.38 & \multicolumn{1}{c|}{12.31} & 14.74 & 7.17 & 11.76 & 11.00 \\
\multicolumn{1}{c|}{\textbf{5}} & 2.93 & 2.54 & 3.05 & \multicolumn{1}{c|}{2.71} & 4.66 & .96 & .07 & \multicolumn{1}{c|}{1.92} & 18.37 & 23.71 & 25.29 & \multicolumn{1}{c|}{13.20} & 30.35 & 21.06 & 22.76 & 6.03 \\
\multicolumn{1}{c|}{\textbf{10}} & 3.09 & 1.97 & 3.40 & \multicolumn{1}{c|}{3.09} & 32.55 & 27.86 & 25.46 & \multicolumn{1}{c|}{18.09} & 31.59 & 28.15 & 30.12 & \multicolumn{1}{c|}{23.83} & 48.88 & 36.44 & 35.56 & 24.74 \\
\multicolumn{1}{c|}{\textbf{20}} & 11.02 & 7.54 & 5.22 & \multicolumn{1}{c|}{3.58} & 47.00 & 36.19 & 27.93 & \multicolumn{1}{c|}{25.06} & 49.58 & 39.37 & 37.10 & \multicolumn{1}{c|}{31.78} & 59.18 & 48.44 & 45.42 & 36.89
\end{tabular}
}
\end{table}

Similar to the heart segmentation in JSRT, the clavicles segmentation followed the same tendency, as showed in table~\ref{tab:jsrt_clavicles}. Although, having lower scores in general, we can see that the methods hold the same order of performance. WeaSeL proving to be a reliable method when the target task is absent in source tasks, or when they highly differ from sources. The overall low scores, can be explained by the use of small resized images, that can deform and blend multiple objects which is problematic for small structures like clavicles.

\section{Visual Examples}

In this section we show some examples of segmentation produced by our method WeaSeL.

Below, in Table~\ref{tab:tasks} is the relation of task and examples. In each example, the lines are the number of shots in the experiment, and the columns are the parameters $n$ for points and $s$ for grids. These parameters, were the ones used to simulate the sparse labels of the support set of the few-shot task.

\begin{table}[h!]
    \centering
    \caption{Relation of examples and tasks}
    \label{tab:tasks}
\begin{tabular}{ccc}
        Dataset &  Class & Figures\\
        \hline
        JSRT &  Lungs & Figures~\ref{tab:jsrt_lungs_visual} and~\ref{tab:jsrt_lungs_visual2}\\
        JSRT &  Clavicles & Figures~\ref{tab:jsrt_clavicles_visual} and~\ref{tab:jsrt_clavicles_visual2}\\
        JSRT &  Heart & Figures~\ref{tab:jsrt_heart_visual} and~\ref{tab:jsrt_heart_visual2}\\
        Montgomery &  Lungs & Figures~\ref{tab:montgomery_lungs_visual} and~\ref{tab:montgomery_lungs_visual2}\\
        OpenIST &  Lungs & Figures~\ref{tab:openist_lungs_visual} and~\ref{tab:openist_lungs_visual2}\\
        Shenzhen &  Lungs & Figures~\ref{tab:shenzhen_lungs_visual} and~\ref{tab:shenzhen_lungs_visual2}\\
    \end{tabular}
    
\end{table}

\newcommand\cincludegraphics[2][]{\raisebox{-0.1\height}{\includegraphics[#1]{#2}}}
\newcommand\exImages{0.15}

\begin{table}[h]

\resizebox{\textwidth}{!}{
\begin{tabular}{cllllllll}
\multicolumn{1}{l}{} & \multicolumn{4}{c}{\textbf{Image}} & \multicolumn{4}{c}{\textbf{Ground Truth}} \\
\multicolumn{1}{l}{} & \multicolumn{4}{c}{
\cincludegraphics[width=\exImages\textwidth]{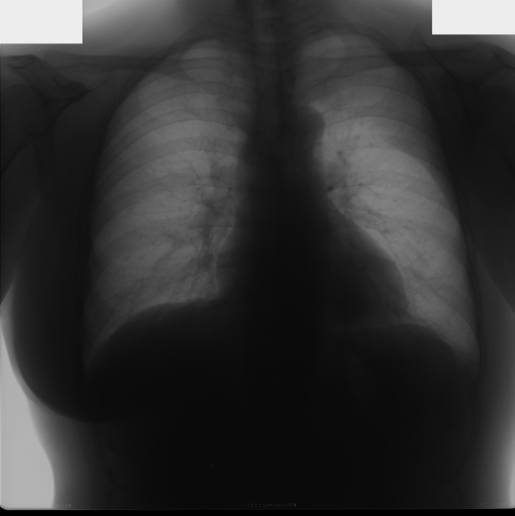}} & \multicolumn{4}{c}{
\cincludegraphics[width=\exImages\textwidth]{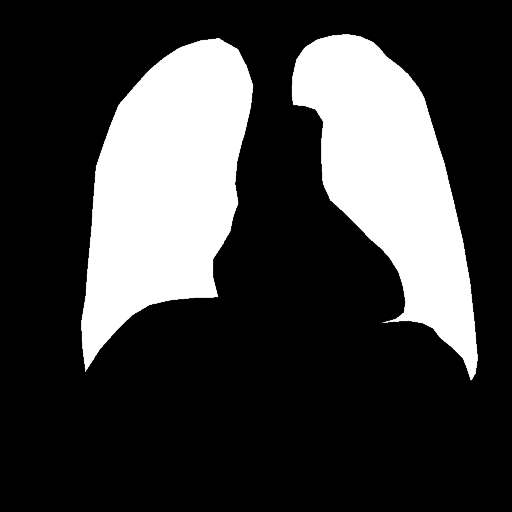}} \\
\multicolumn{9}{l}{} \\
\multicolumn{1}{l}{} & \multicolumn{4}{c}{\textbf{Points}} & \multicolumn{4}{c}{\textbf{Grid}} \\
\textbf{Shots} & \multicolumn{1}{c}{\textbf{1}} & \multicolumn{1}{c}{\textbf{5}} & \multicolumn{1}{c}{\textbf{10}} & \multicolumn{1}{c}{\textbf{20}} & \multicolumn{1}{c}{\textbf{8}} & \multicolumn{1}{c}{\textbf{12}} & \multicolumn{1}{c}{\textbf{16}} & \multicolumn{1}{c}{\textbf{20}} \\
\textbf{1} & 
\cincludegraphics[width=\exImages\textwidth]{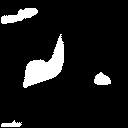} &
\cincludegraphics[width=\exImages\textwidth]{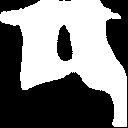} & 
\cincludegraphics[width=\exImages\textwidth]{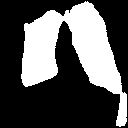}& 
\cincludegraphics[width=\exImages\textwidth]{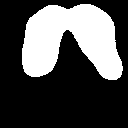}&
\cincludegraphics[width=\exImages\textwidth]{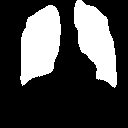} &
\cincludegraphics[width=\exImages\textwidth]{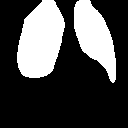} & 
\cincludegraphics[width=\exImages\textwidth]{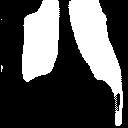}& 
\cincludegraphics[width=\exImages\textwidth]{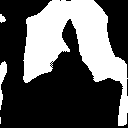}\\
\end{tabular}
}
\captionof{figure}{Example 1, Visual Examples of Segmentation from JSRT Lungs task}
\label{tab:jsrt_lungs_visual}
\end{table}

\begin{table}[h]

\resizebox{\textwidth}{!}{
\begin{tabular}{cllllllll}
\multicolumn{1}{l}{} & \multicolumn{4}{c}{\textbf{Image}} & \multicolumn{4}{c}{\textbf{Ground Truth}} \\
\multicolumn{1}{l}{} & \multicolumn{4}{c}{
\cincludegraphics[width=\exImages\textwidth]{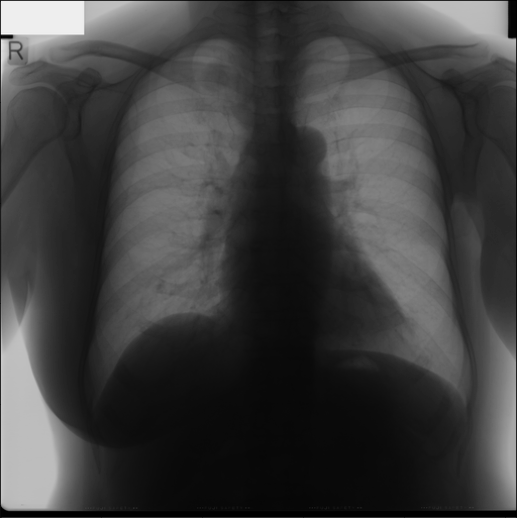}} & \multicolumn{4}{c}{
\cincludegraphics[width=\exImages\textwidth]{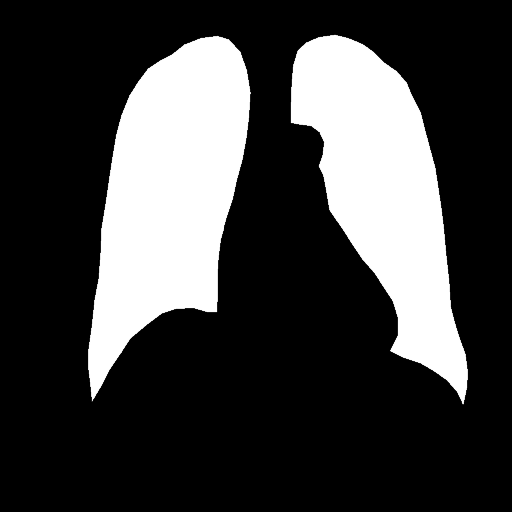}} \\
\multicolumn{9}{l}{} \\
\multicolumn{1}{l}{} & \multicolumn{4}{c}{\textbf{Points}} & \multicolumn{4}{c}{\textbf{Grid}} \\
\textbf{Shots} & \multicolumn{1}{c}{\textbf{1}} & \multicolumn{1}{c}{\textbf{5}} & \multicolumn{1}{c}{\textbf{10}} & \multicolumn{1}{c}{\textbf{20}} & \multicolumn{1}{c}{\textbf{8}} & \multicolumn{1}{c}{\textbf{12}} & \multicolumn{1}{c}{\textbf{16}} & \multicolumn{1}{c}{\textbf{20}} \\
\textbf{1} & 
\cincludegraphics[width=\exImages\textwidth]{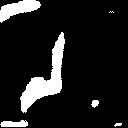} &
\cincludegraphics[width=\exImages\textwidth]{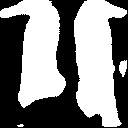} & 
\cincludegraphics[width=\exImages\textwidth]{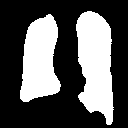}& 
\cincludegraphics[width=\exImages\textwidth]{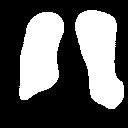}&
\cincludegraphics[width=\exImages\textwidth]{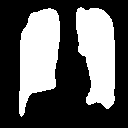} &
\cincludegraphics[width=\exImages\textwidth]{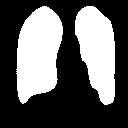} & 
\cincludegraphics[width=\exImages\textwidth]{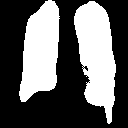}& 
\cincludegraphics[width=\exImages\textwidth]{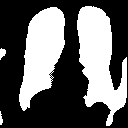}\\
\end{tabular}
}
\captionof{figure}{Example 2, Visual Examples of Segmentation from JSRT Lungs task}
\label{tab:jsrt_lungs_visual2}
\end{table}

\begin{table}[h]

\resizebox{\textwidth}{!}{
\begin{tabular}{cllllllll}
\multicolumn{1}{l}{} & \multicolumn{4}{c}{\textbf{Image}} & \multicolumn{4}{c}{\textbf{Ground Truth}} \\
\multicolumn{1}{l}{} & \multicolumn{4}{c}{
\cincludegraphics[width=\exImages\textwidth]{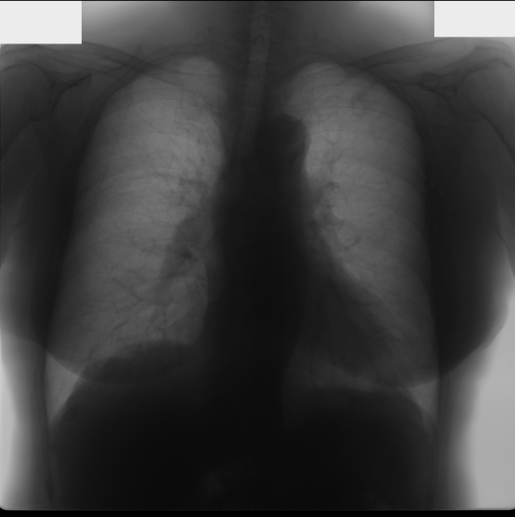}} & \multicolumn{4}{c}{
\cincludegraphics[width=\exImages\textwidth]{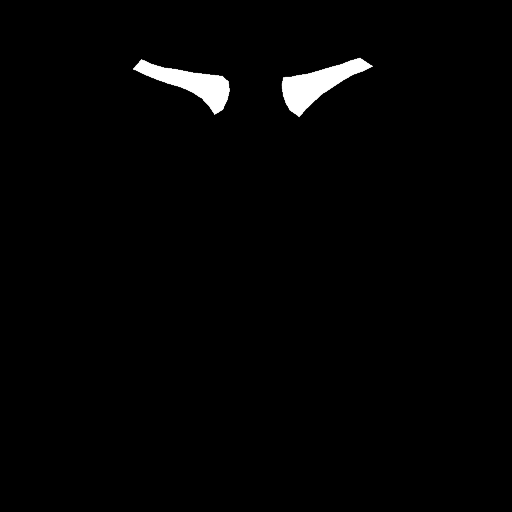}} \\
\multicolumn{9}{l}{} \\
\multicolumn{1}{l}{} & \multicolumn{4}{c}{\textbf{Points}} & \multicolumn{4}{c}{\textbf{Grid}} \\
\textbf{Shots} & \multicolumn{1}{c}{\textbf{1}} & \multicolumn{1}{c}{\textbf{5}} & \multicolumn{1}{c}{\textbf{10}} & \multicolumn{1}{c}{\textbf{20}} & \multicolumn{1}{c}{\textbf{8}} & \multicolumn{1}{c}{\textbf{12}} & \multicolumn{1}{c}{\textbf{16}} & \multicolumn{1}{c}{\textbf{20}} \\
\textbf{1} & 
\cincludegraphics[width=\exImages\textwidth]{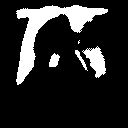} &
\cincludegraphics[width=\exImages\textwidth]{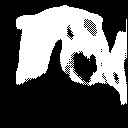} & 
\cincludegraphics[width=\exImages\textwidth]{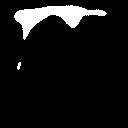}& 
\cincludegraphics[width=\exImages\textwidth]{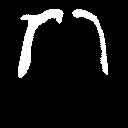}&
\cincludegraphics[width=\exImages\textwidth]{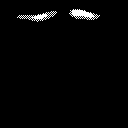} &
\cincludegraphics[width=\exImages\textwidth]{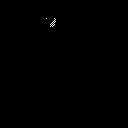} & 
\cincludegraphics[width=\exImages\textwidth]{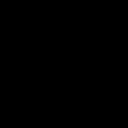}& 
\cincludegraphics[width=\exImages\textwidth]{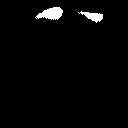}\\
\end{tabular}
}
\captionof{figure}{Example 3, Visual Examples of Segmentation from JSRT Clavicles task}
\label{tab:jsrt_clavicles_visual}
\end{table}

\begin{table}[h]

\resizebox{\textwidth}{!}{
\begin{tabular}{cllllllll}
\multicolumn{1}{l}{} & \multicolumn{4}{c}{\textbf{Image}} & \multicolumn{4}{c}{\textbf{Ground Truth}} \\
\multicolumn{1}{l}{} & \multicolumn{4}{c}{
\cincludegraphics[width=\exImages\textwidth]{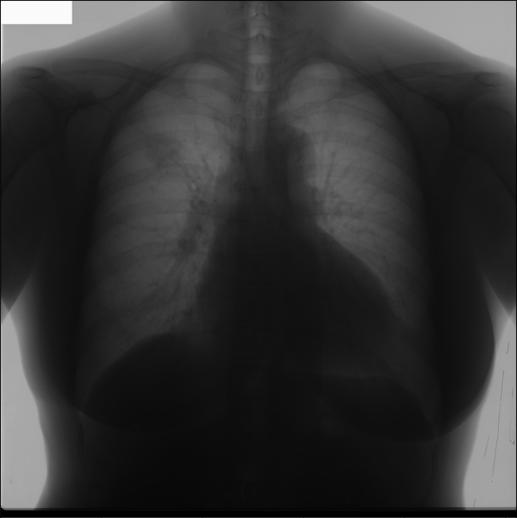}} & \multicolumn{4}{c}{
\cincludegraphics[width=\exImages\textwidth]{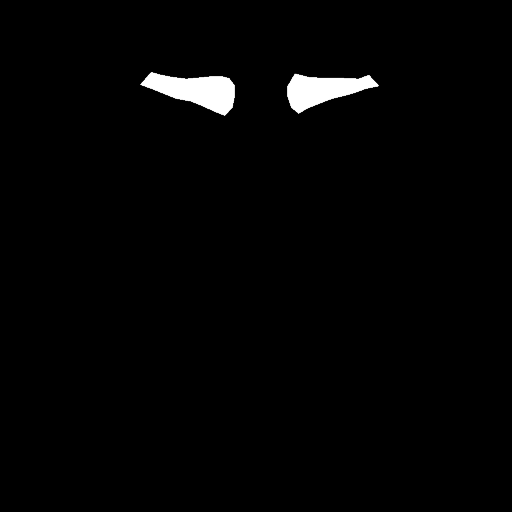}} \\
\multicolumn{9}{l}{} \\
\multicolumn{1}{l}{} & \multicolumn{4}{c}{\textbf{Points}} & \multicolumn{4}{c}{\textbf{Grid}} \\
\textbf{Shots} & \multicolumn{1}{c}{\textbf{1}} & \multicolumn{1}{c}{\textbf{5}} & \multicolumn{1}{c}{\textbf{10}} & \multicolumn{1}{c}{\textbf{20}} & \multicolumn{1}{c}{\textbf{8}} & \multicolumn{1}{c}{\textbf{12}} & \multicolumn{1}{c}{\textbf{16}} & \multicolumn{1}{c}{\textbf{20}} \\
\textbf{1} & 
\cincludegraphics[width=\exImages\textwidth]{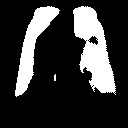} &
\cincludegraphics[width=\exImages\textwidth]{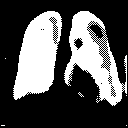} & 
\cincludegraphics[width=\exImages\textwidth]{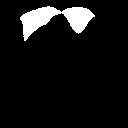}& 
\cincludegraphics[width=\exImages\textwidth]{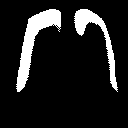}&
\cincludegraphics[width=\exImages\textwidth]{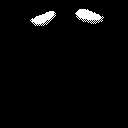} &
\cincludegraphics[width=\exImages\textwidth]{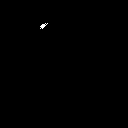} & 
\cincludegraphics[width=\exImages\textwidth]{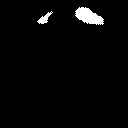}& 
\cincludegraphics[width=\exImages\textwidth]{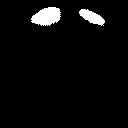}\\
\end{tabular}
}
\captionof{figure}{Example 4, Visual Examples of Segmentation from JSRT Clavicles task}
\label{tab:jsrt_clavicles_visual2}
\end{table}

\begin{table}[h]

\resizebox{\textwidth}{!}{
\begin{tabular}{cllllllll}
\multicolumn{1}{l}{} & \multicolumn{4}{c}{\textbf{Image}} & \multicolumn{4}{c}{\textbf{Ground Truth}} \\
\multicolumn{1}{l}{} & \multicolumn{4}{c}{
\cincludegraphics[width=\exImages\textwidth]{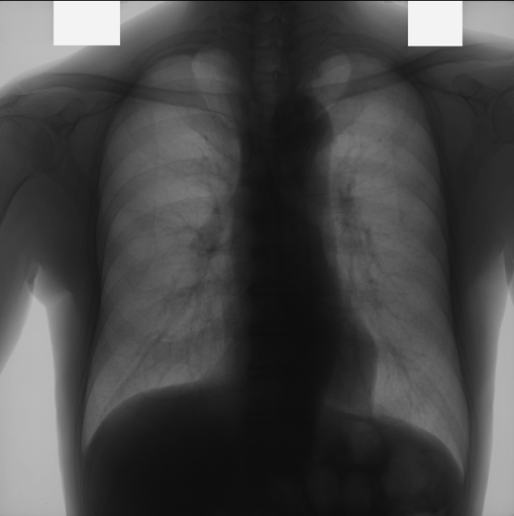}} & \multicolumn{4}{c}{
\cincludegraphics[width=\exImages\textwidth]{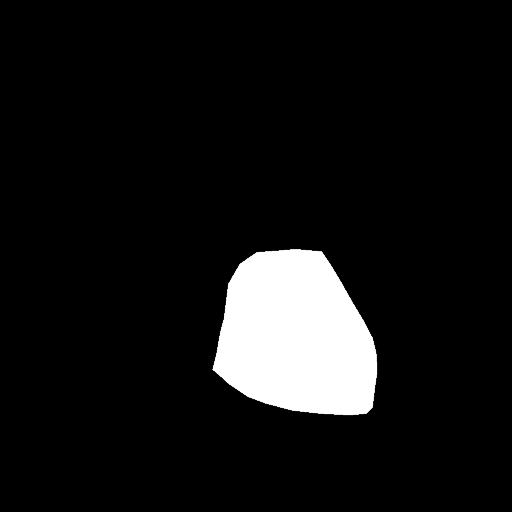}} \\
\multicolumn{9}{l}{} \\
\multicolumn{1}{l}{} & \multicolumn{4}{c}{\textbf{Points}} & \multicolumn{4}{c}{\textbf{Grid}} \\
\textbf{Shots} & \multicolumn{1}{c}{\textbf{1}} & \multicolumn{1}{c}{\textbf{5}} & \multicolumn{1}{c}{\textbf{10}} & \multicolumn{1}{c}{\textbf{20}} & \multicolumn{1}{c}{\textbf{8}} & \multicolumn{1}{c}{\textbf{12}} & \multicolumn{1}{c}{\textbf{16}} & \multicolumn{1}{c}{\textbf{20}} \\
\textbf{1} & 
\cincludegraphics[width=\exImages\textwidth]{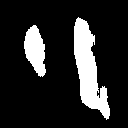} &
\cincludegraphics[width=\exImages\textwidth]{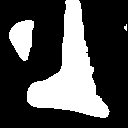} & 
\cincludegraphics[width=\exImages\textwidth]{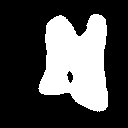}& 
\cincludegraphics[width=\exImages\textwidth]{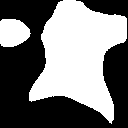}&
\cincludegraphics[width=\exImages\textwidth]{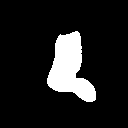} &
\cincludegraphics[width=\exImages\textwidth]{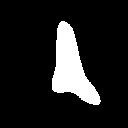} & 
\cincludegraphics[width=\exImages\textwidth]{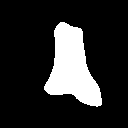}& 
\cincludegraphics[width=\exImages\textwidth]{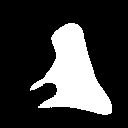}\\
\end{tabular}
}
\captionof{figure}{Example 5, Visual Examples of Segmentation from JSRT Heart task}
\label{tab:jsrt_heart_visual}
\end{table}

\begin{table}[h]

\resizebox{\textwidth}{!}{
\begin{tabular}{cllllllll}
\multicolumn{1}{l}{} & \multicolumn{4}{c}{\textbf{Image}} & \multicolumn{4}{c}{\textbf{Ground Truth}} \\
\multicolumn{1}{l}{} & \multicolumn{4}{c}{
\cincludegraphics[width=\exImages\textwidth]{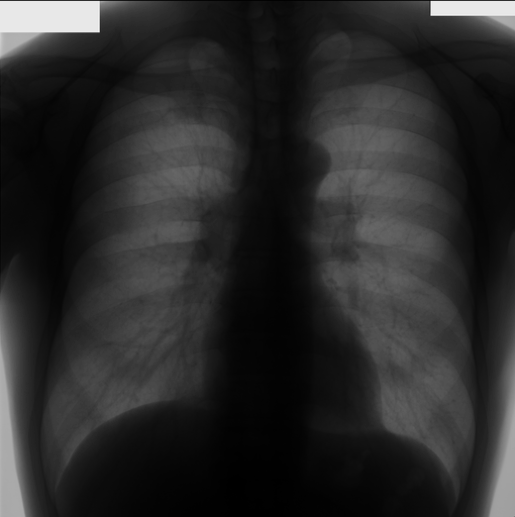}} & \multicolumn{4}{c}{
\cincludegraphics[width=\exImages\textwidth]{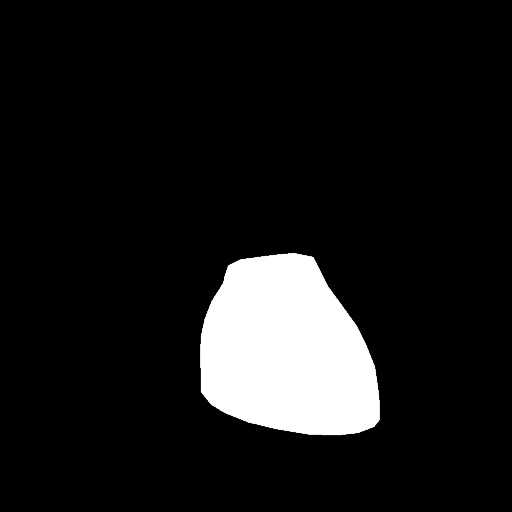}} \\
\multicolumn{9}{l}{} \\
\multicolumn{1}{l}{} & \multicolumn{4}{c}{\textbf{Points}} & \multicolumn{4}{c}{\textbf{Grid}} \\
\textbf{Shots} & \multicolumn{1}{c}{\textbf{1}} & \multicolumn{1}{c}{\textbf{5}} & \multicolumn{1}{c}{\textbf{10}} & \multicolumn{1}{c}{\textbf{20}} & \multicolumn{1}{c}{\textbf{8}} & \multicolumn{1}{c}{\textbf{12}} & \multicolumn{1}{c}{\textbf{16}} & \multicolumn{1}{c}{\textbf{20}} \\
\textbf{1} & 
\cincludegraphics[width=\exImages\textwidth]{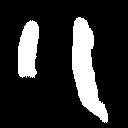} &
\cincludegraphics[width=\exImages\textwidth]{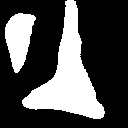} & 
\cincludegraphics[width=\exImages\textwidth]{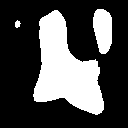}& 
\cincludegraphics[width=\exImages\textwidth]{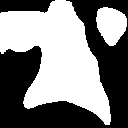}&
\cincludegraphics[width=\exImages\textwidth]{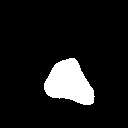} &
\cincludegraphics[width=\exImages\textwidth]{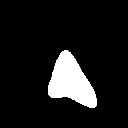} & 
\cincludegraphics[width=\exImages\textwidth]{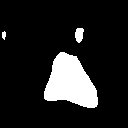}& 
\cincludegraphics[width=\exImages\textwidth]{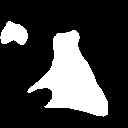}\\
\end{tabular}
}
\captionof{figure}{Example 6, Visual Examples of Segmentation from JSRT Heart task}
\label{tab:jsrt_heart_visual2}
\end{table}

\begin{table}[h]

\resizebox{\textwidth}{!}{
\begin{tabular}{cllllllll}
\multicolumn{1}{l}{} & \multicolumn{4}{c}{\textbf{Image}} & \multicolumn{4}{c}{\textbf{Ground Truth}} \\
\multicolumn{1}{l}{} & \multicolumn{4}{c}{
\cincludegraphics[width=\exImages\textwidth]{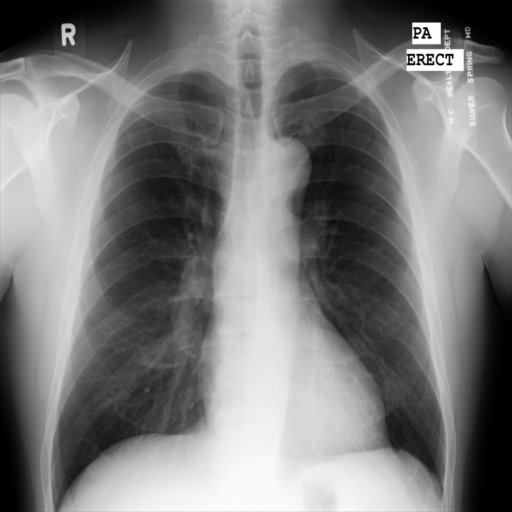}} & \multicolumn{4}{c}{
\cincludegraphics[width=\exImages\textwidth]{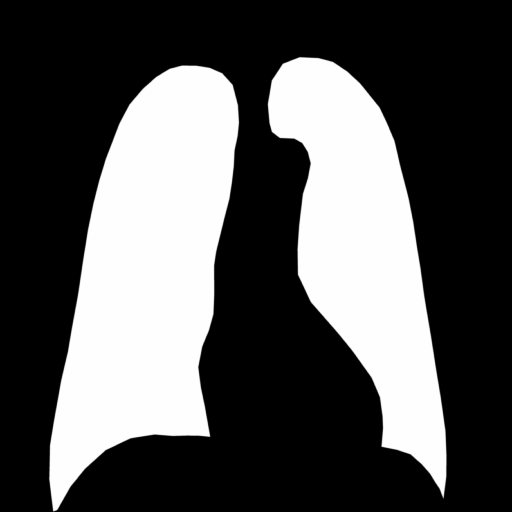}} \\
\multicolumn{9}{l}{} \\
\multicolumn{1}{l}{} & \multicolumn{4}{c}{\textbf{Points}} & \multicolumn{4}{c}{\textbf{Grid}} \\
\textbf{Shots} & \multicolumn{1}{c}{\textbf{1}} & \multicolumn{1}{c}{\textbf{5}} & \multicolumn{1}{c}{\textbf{10}} & \multicolumn{1}{c}{\textbf{20}} & \multicolumn{1}{c}{\textbf{8}} & \multicolumn{1}{c}{\textbf{12}} & \multicolumn{1}{c}{\textbf{16}} & \multicolumn{1}{c}{\textbf{20}} \\
\textbf{1} & 
\cincludegraphics[width=\exImages\textwidth]{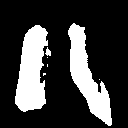} &
\cincludegraphics[width=\exImages\textwidth]{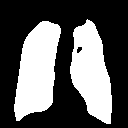} & 
\cincludegraphics[width=\exImages\textwidth]{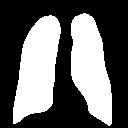}& 
\cincludegraphics[width=\exImages\textwidth]{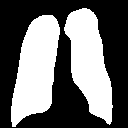}&
\cincludegraphics[width=\exImages\textwidth]{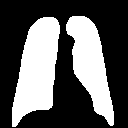} &
\cincludegraphics[width=\exImages\textwidth]{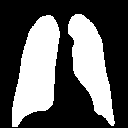} & 
\cincludegraphics[width=\exImages\textwidth]{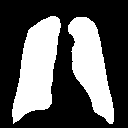}& 
\cincludegraphics[width=\exImages\textwidth]{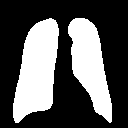}\\
\end{tabular}
}
\captionof{figure}{Example 7, Visual Examples of Segmentation from Montgomery Lungs task}
\label{tab:montgomery_lungs_visual}
\end{table}

\begin{table}[h]

\resizebox{\textwidth}{!}{
\begin{tabular}{cllllllll}
\multicolumn{1}{l}{} & \multicolumn{4}{c}{\textbf{Image}} & \multicolumn{4}{c}{\textbf{Ground Truth}} \\
\multicolumn{1}{l}{} & \multicolumn{4}{c}{
\cincludegraphics[width=\exImages\textwidth]{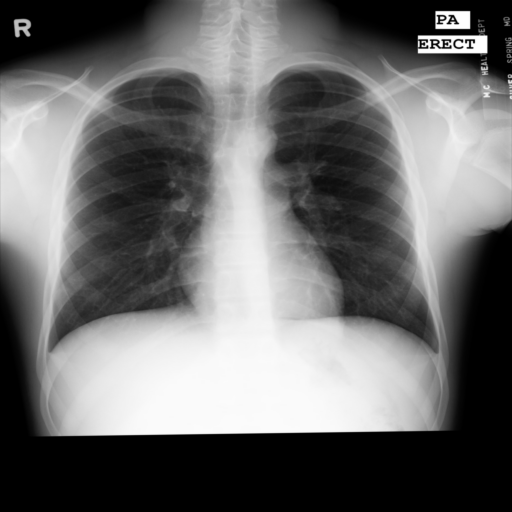}} & \multicolumn{4}{c}{
\cincludegraphics[width=\exImages\textwidth]{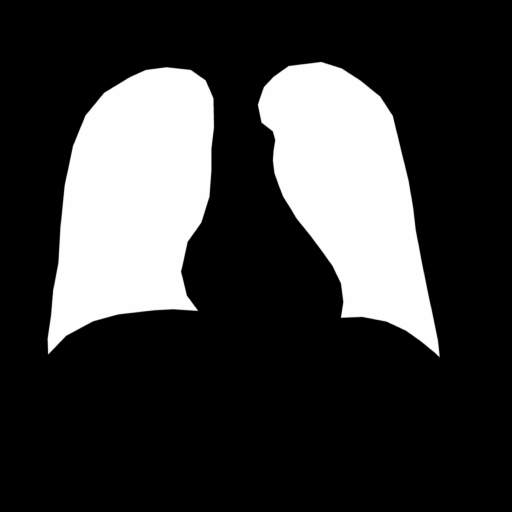}} \\
\multicolumn{9}{l}{} \\
\multicolumn{1}{l}{} & \multicolumn{4}{c}{\textbf{Points}} & \multicolumn{4}{c}{\textbf{Grid}} \\
\textbf{Shots} & \multicolumn{1}{c}{\textbf{1}} & \multicolumn{1}{c}{\textbf{5}} & \multicolumn{1}{c}{\textbf{10}} & \multicolumn{1}{c}{\textbf{20}} & \multicolumn{1}{c}{\textbf{8}} & \multicolumn{1}{c}{\textbf{12}} & \multicolumn{1}{c}{\textbf{16}} & \multicolumn{1}{c}{\textbf{20}} \\
\textbf{1} & 
\cincludegraphics[width=\exImages\textwidth]{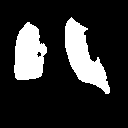} &
\cincludegraphics[width=\exImages\textwidth]{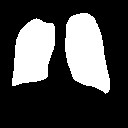} & 
\cincludegraphics[width=\exImages\textwidth]{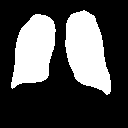}& 
\cincludegraphics[width=\exImages\textwidth]{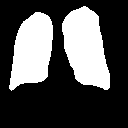}&
\cincludegraphics[width=\exImages\textwidth]{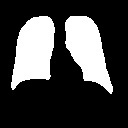} &
\cincludegraphics[width=\exImages\textwidth]{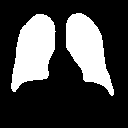} & 
\cincludegraphics[width=\exImages\textwidth]{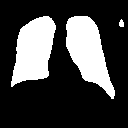}& 
\cincludegraphics[width=\exImages\textwidth]{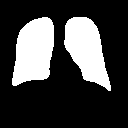}\\
\end{tabular}
}
\captionof{figure}{Example 8, Visual Examples of Segmentation from Montgomery Lungs task}
\label{tab:montgomery_lungs_visual2}
\end{table}

\begin{table}[h]

\resizebox{\textwidth}{!}{
\begin{tabular}{cllllllll}
\multicolumn{1}{l}{} & \multicolumn{4}{c}{\textbf{Image}} & \multicolumn{4}{c}{\textbf{Ground Truth}} \\
\multicolumn{1}{l}{} & \multicolumn{4}{c}{
\cincludegraphics[width=\exImages\textwidth]{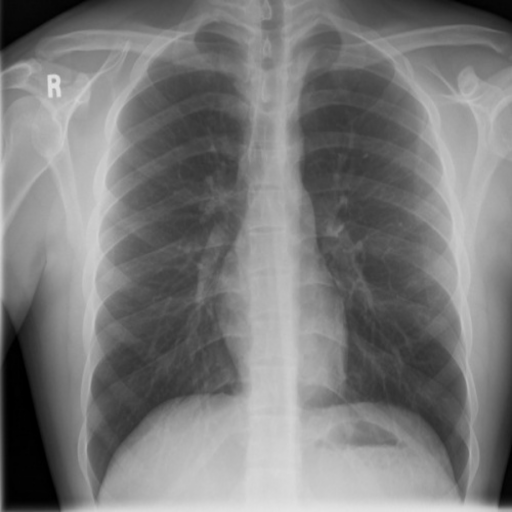}} & \multicolumn{4}{c}{
\cincludegraphics[width=\exImages\textwidth]{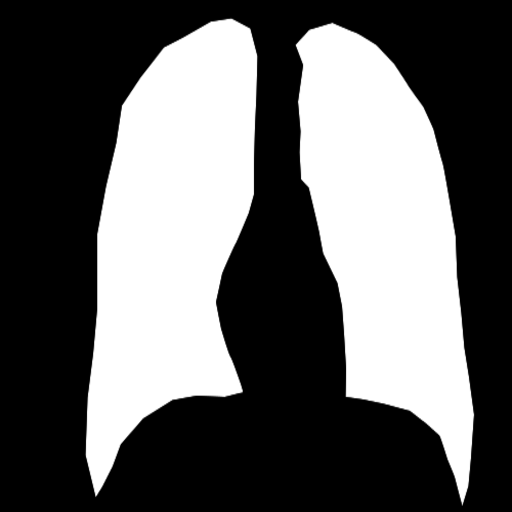}} \\
\multicolumn{9}{l}{} \\
\multicolumn{1}{l}{} & \multicolumn{4}{c}{\textbf{Points}} & \multicolumn{4}{c}{\textbf{Grid}} \\
\textbf{Shots} & \multicolumn{1}{c}{\textbf{1}} & \multicolumn{1}{c}{\textbf{5}} & \multicolumn{1}{c}{\textbf{10}} & \multicolumn{1}{c}{\textbf{20}} & \multicolumn{1}{c}{\textbf{8}} & \multicolumn{1}{c}{\textbf{12}} & \multicolumn{1}{c}{\textbf{16}} & \multicolumn{1}{c}{\textbf{20}} \\
\textbf{1} & 
\cincludegraphics[width=\exImages\textwidth]{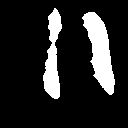} &
\cincludegraphics[width=\exImages\textwidth]{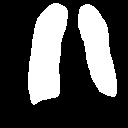} & 
\cincludegraphics[width=\exImages\textwidth]{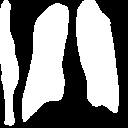}& 
\cincludegraphics[width=\exImages\textwidth]{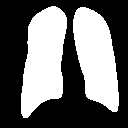}&
\cincludegraphics[width=\exImages\textwidth]{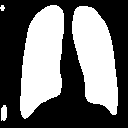} &
\cincludegraphics[width=\exImages\textwidth]{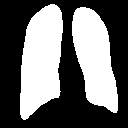} & 
\cincludegraphics[width=\exImages\textwidth]{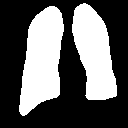}& 
\cincludegraphics[width=\exImages\textwidth]{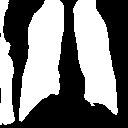}\\
\end{tabular}
}
\captionof{figure}{Example 9, Visual Examples of Segmentation from OpenIST Lungs task}
\label{tab:openist_lungs_visual}
\end{table}

\begin{table}[h]

\resizebox{\textwidth}{!}{
\begin{tabular}{cllllllll}
\multicolumn{1}{l}{} & \multicolumn{4}{c}{\textbf{Image}} & \multicolumn{4}{c}{\textbf{Ground Truth}} \\
\multicolumn{1}{l}{} & \multicolumn{4}{c}{
\cincludegraphics[width=\exImages\textwidth]{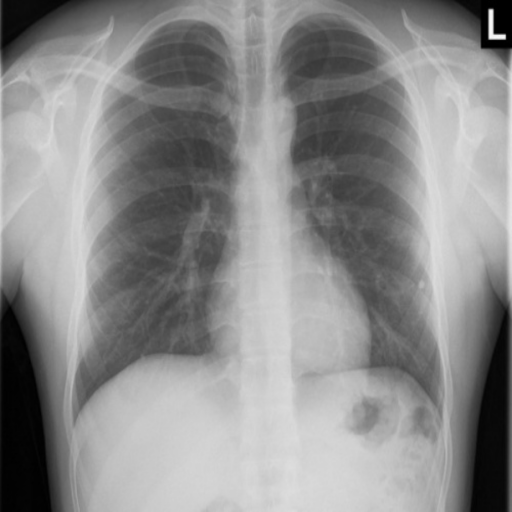}} & \multicolumn{4}{c}{
\cincludegraphics[width=\exImages\textwidth]{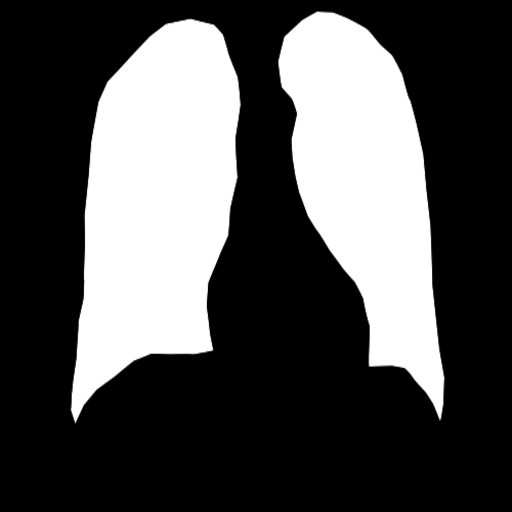}} \\
\multicolumn{9}{l}{} \\
\multicolumn{1}{l}{} & \multicolumn{4}{c}{\textbf{Points}} & \multicolumn{4}{c}{\textbf{Grid}} \\
\textbf{Shots} & \multicolumn{1}{c}{\textbf{1}} & \multicolumn{1}{c}{\textbf{5}} & \multicolumn{1}{c}{\textbf{10}} & \multicolumn{1}{c}{\textbf{20}} & \multicolumn{1}{c}{\textbf{8}} & \multicolumn{1}{c}{\textbf{12}} & \multicolumn{1}{c}{\textbf{16}} & \multicolumn{1}{c}{\textbf{20}} \\
\textbf{1} & 
\cincludegraphics[width=\exImages\textwidth]{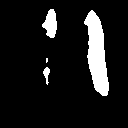} &
\cincludegraphics[width=\exImages\textwidth]{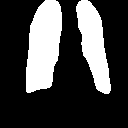} & 
\cincludegraphics[width=\exImages\textwidth]{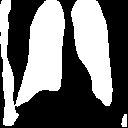}& 
\cincludegraphics[width=\exImages\textwidth]{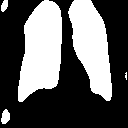}&
\cincludegraphics[width=\exImages\textwidth]{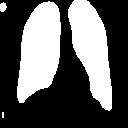} &
\cincludegraphics[width=\exImages\textwidth]{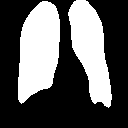} & 
\cincludegraphics[width=\exImages\textwidth]{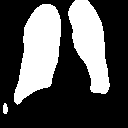}& 
\cincludegraphics[width=\exImages\textwidth]{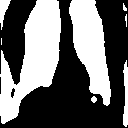}\\
\end{tabular}
}
\captionof{figure}{Example 10, Visual Examples of Segmentation from OpenIST Lungs task}
\label{tab:openist_lungs_visual2}
\end{table}

\begin{table}[h]

\resizebox{\textwidth}{!}{
\begin{tabular}{cllllllll}
\multicolumn{1}{l}{} & \multicolumn{4}{c}{\textbf{Image}} & \multicolumn{4}{c}{\textbf{Ground Truth}} \\
\multicolumn{1}{l}{} & \multicolumn{4}{c}{
\cincludegraphics[width=\exImages\textwidth]{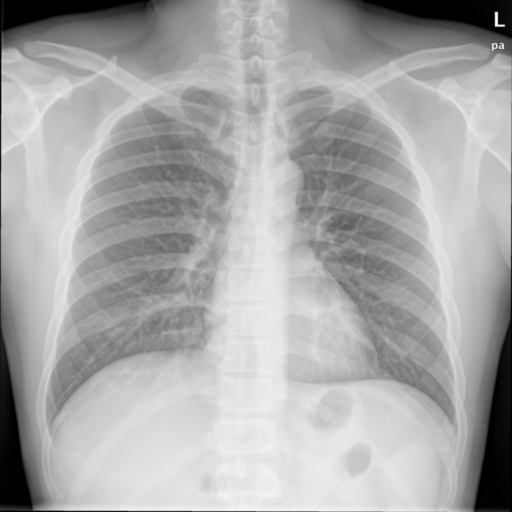}} & \multicolumn{4}{c}{
\cincludegraphics[width=\exImages\textwidth]{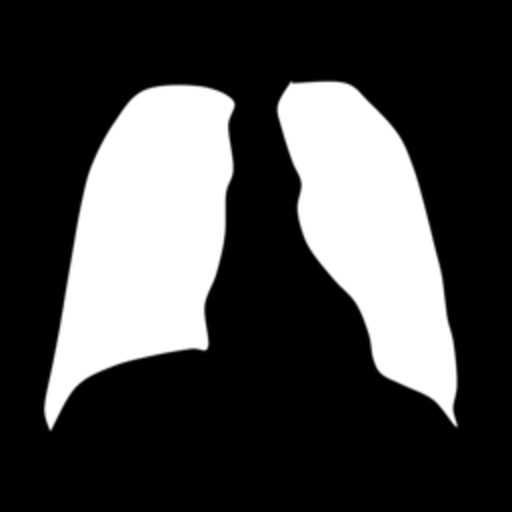}} \\
\multicolumn{9}{l}{} \\
\multicolumn{1}{l}{} & \multicolumn{4}{c}{\textbf{Points}} & \multicolumn{4}{c}{\textbf{Grid}} \\
\textbf{Shots} & \multicolumn{1}{c}{\textbf{1}} & \multicolumn{1}{c}{\textbf{5}} & \multicolumn{1}{c}{\textbf{10}} & \multicolumn{1}{c}{\textbf{20}} & \multicolumn{1}{c}{\textbf{8}} & \multicolumn{1}{c}{\textbf{12}} & \multicolumn{1}{c}{\textbf{16}} & \multicolumn{1}{c}{\textbf{20}} \\
\textbf{1} & 
\cincludegraphics[width=\exImages\textwidth]{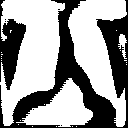} &
\cincludegraphics[width=\exImages\textwidth]{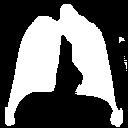} & 
\cincludegraphics[width=\exImages\textwidth]{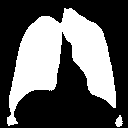}& 
\cincludegraphics[width=\exImages\textwidth]{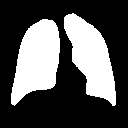}&
\cincludegraphics[width=\exImages\textwidth]{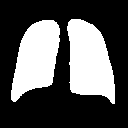} &
\cincludegraphics[width=\exImages\textwidth]{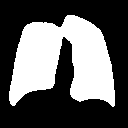} & 
\cincludegraphics[width=\exImages\textwidth]{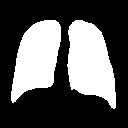}& 
\cincludegraphics[width=\exImages\textwidth]{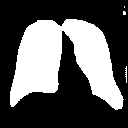}\\
\end{tabular}
}
\captionof{figure}{Example 11, Visual Examples of Segmentation from Shenzhen Lungs task}
\label{tab:shenzhen_lungs_visual}
\end{table}

\begin{table}[h]

\resizebox{\textwidth}{!}{
\begin{tabular}{cllllllll}
\multicolumn{1}{l}{} & \multicolumn{4}{c}{\textbf{Image}} & \multicolumn{4}{c}{\textbf{Ground Truth}} \\
\multicolumn{1}{l}{} & \multicolumn{4}{c}{
\cincludegraphics[width=\exImages\textwidth]{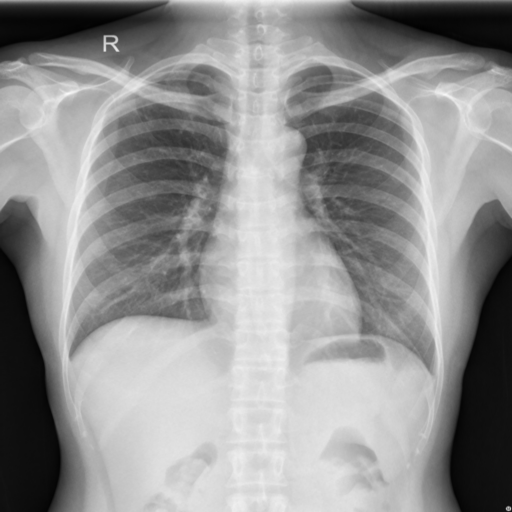}} & \multicolumn{4}{c}{
\cincludegraphics[width=\exImages\textwidth]{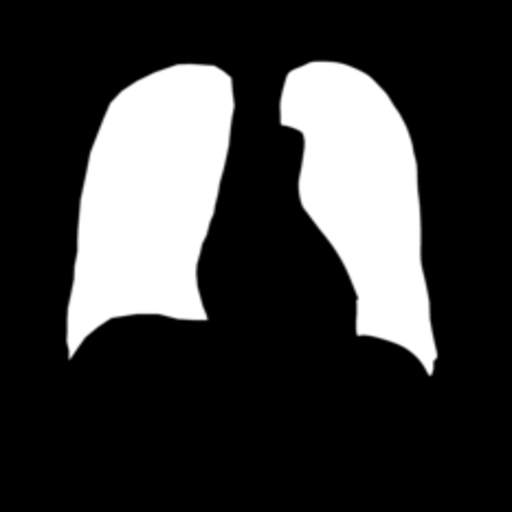}} \\
\multicolumn{9}{l}{} \\
\multicolumn{1}{l}{} & \multicolumn{4}{c}{\textbf{Points}} & \multicolumn{4}{c}{\textbf{Grid}} \\
\textbf{Shots} & \multicolumn{1}{c}{\textbf{1}} & \multicolumn{1}{c}{\textbf{5}} & \multicolumn{1}{c}{\textbf{10}} & \multicolumn{1}{c}{\textbf{20}} & \multicolumn{1}{c}{\textbf{8}} & \multicolumn{1}{c}{\textbf{12}} & \multicolumn{1}{c}{\textbf{16}} & \multicolumn{1}{c}{\textbf{20}} \\
\textbf{1} & 
\cincludegraphics[width=\exImages\textwidth]{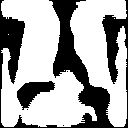} &
\cincludegraphics[width=\exImages\textwidth]{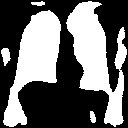} & 
\cincludegraphics[width=\exImages\textwidth]{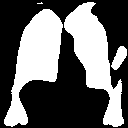}& 
\cincludegraphics[width=\exImages\textwidth]{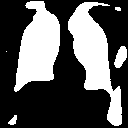}&
\cincludegraphics[width=\exImages\textwidth]{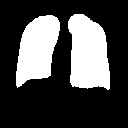} &
\cincludegraphics[width=\exImages\textwidth]{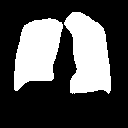} & 
\cincludegraphics[width=\exImages\textwidth]{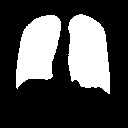}& 
\cincludegraphics[width=\exImages\textwidth]{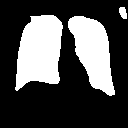}\\
\end{tabular}
}
\captionof{figure}{Example 12, Visual Examples of Segmentation from Shenzhen Lungs task}
\label{tab:shenzhen_lungs_visual2}
\end{table}